\documentclass[journal ]{new-aiaa}
\usepackage[utf8]{inputenc}
\usepackage{textcomp}

\usepackage{graphicx}
\usepackage{amsmath}
\usepackage[version=4]{mhchem}
\usepackage{siunitx}
\usepackage{longtable,tabularx}
\usepackage{relsize}
\usepackage{acronym}
\usepackage{mathtools}
\usepackage{subcaption}
\usepackage{caption}
\usepackage{xfrac}

\newtheorem{lemma}{Lemma}

\newacro{SCvx}{Successive Convexification}
\newacro{SCP}{Sequential Convex Programming}
\newacro{DRO}{Distant Retrograde Orbit}
\newacro{NRHO}{Near-Rectilinear Halo Orbit}
\newacro{CRTBP}{Circular Restricted Three Body Problem}
\newacro{CRLB}{Cramer-Rao Lower Bound}
\newacro{RMS}{Root Mean Square}
\newacro{SSA}{Space Situational Awareness}
\newacro{SDA}{Space Domain Awareness}
\newacro{DSN}{Deep Space Network}
\newacro{SSAT}{Spacecraft-to-Spacecraft Absolute Tracking}
\newacro{PSD}{Power Spectral Density}
\newacro{IID}{Independent and Identically Distributed}
\newacro{KLD}{Kullback-Leibler Divergence}
\newacro{RSS}{Root Sum Square}
\newacro{SO}{Space Object}
\newacro{GEO}{Geostationary Orbit}
\newacro{LEO}{Low Earth Orbit}
\newacro{GNC}{Guidance, Navigation, and Control}
\newacro{NASA}{National Aeronautics and Space Administration}
\newacro{AFRL}{Air Force Research Laboratory}
\newacro{CAPSTONE}{Cislunar Autonomous Positioning System Technology Operations and Navigation Experiment}
\newacro{PDF}{Probability Density Function}
\newacro{DCP}{Disciplined Convex Programming}
\newacro{ODE}{Ordinary Differential Equation}
\newacro{JPL}{Jet Propulsion Laboratory}

\title{Information-Based Trajectory Planning for Spacecraft-to-Spacecraft Tracking and Navigation in Cislunar Space}

\author{Trevor N. Wolf \footnote{Postdoctoral Research Associate, Colorado Center for Astrodynamics Research.} and Jay W. McMahon\footnote{Associate Professor, Colorado Center for Astrodynamics Research.}}
\affil{University of Colorado Boulder, Boulder, CO, 80309}
\author{Brandon A. Jones \footnote{Associate Professor, Department of Aerospace Engineering and Engineering Mechanics.}}
\affil{The University of Texas at Austin, Austin, TX, 78712}

\begin{document}

\maketitle

\begin{abstract}

We present a trajectory planning method that balances information collection and control effort to improve cislunar spacecraft-to-spacecraft absolute tracking. Expanding use of cislunar space requires alternative navigation and tracking procedures that minimize reliance on ground-based support. Among efforts to address this need, spacecraft-to-spacecraft tracking exploits nonlinear dynamical tracers encoded in a series of relative measurements to infer absolute states of both an observer and a target. The geometry between spacecraft operating under this mode can significantly influence tracking performance. This work considers this geometrical impact by designing observer trajectories that jointly balance control effort and an information-theoretic quantification of the expected spacecraft-to-spacecraft tracking performance. Our methods are designed for multiple low-thrust observation platforms of various sensing modalities and can incorporate multiple space object targets in planning. By leveraging information gain in the optimal control problem, we report almost an order of magnitude improvement in the expected navigation and tracking errors for an optical observer operating in a Distant Retrograde Orbit (DRO). This work demonstrates the feasibility and value of information-optimal low-thrust spacecraft trajectory design for current and upcoming cislunar missions. 

\end{abstract}

%
\section{Introduction}\label{sec:introduction}
\subsection{Motivation and Background}

The expanding scope of cislunar spaceflight requires advanced tracking, navigation, and planning algorithms that address challenges not encountered in near-Earth operations. The vast volume of the Earth-Moon dynamical region, lack of geometric diversity over observation arcs, lunar occultations, long baseline distances, and high-contrast optical observations, limit efficient surveillance with Earth-based \ac{SO} tracking facilities \cite{Frueh_2021, Holzinger_2021, Bhadauria_2022, Baker_2024}. Placing observation assets directly in cislunar space, such as the experiments proposed in Ref. \cite{Oracle_2022}, can ameliorate these issues, yet introduce new operational complexities. Nonlinear gravitational interactions between spacecraft and the Earth, Moon, and Sun in this region complicate accurate uncertainty quantification -- essential for navigation, safe trajectory planning, and sensor management -- particularly over extended time intervals \cite{Wolf_2021, Jones_2024, Reifler_2024}. Additionally, accelerated cislunar operations will outpace the capacity of navigation relays, such as the \ac{DSN}, that have, so far, been necessary for cislunar spaceflight. Minimizing reliance on Earth-based resources, while meeting operational constraints, requires developing autonomous cislunar \ac{GNC} technologies. Notably, NASA's Lunar Gateway program requires the station to operate up to 21 days without ground-based support \cite{Badger_2024}. Among efforts to automate cislunar \ac{GNC} and surveillance are systems based on inter-satellite relative measurements.

Markley was the first to propose using inertially referenced relative position measurements in conjunction with geographic landmarks for achieving absolute navigation of two spacecraft in Earth orbit \cite{Markley_1984}. Subsequent studies investigate similar navigation systems, in the absence of landmark features, for satellite constellations (e.g., \cite{Menn_1986, Herklotz_1988, Ananda_1990, Psiaki_1999}).  Later research built upon this concept for gravity determination \cite{Psiaki_2011, Leonard_2012}, absolute navigation in cislunar space through crosslink range measurements \cite{Hill_2008, Wang_2019}, and optical measurements \cite{Greaves_2021}. Indeed, peer-to-peer radiometric autonomous navigation was experimentally validated for cislunar spaceflight with the \ac{CAPSTONE} \cite{Cheetham_2021}. Collectively, this form of navigation has been coined \ac{SSAT}, and we adopt this term, regardless of the observation mode (i.e., bearing angles, ranging, relative position).


\subsection{Proposed Approach}

The relative geometry of spacecraft operating under this archetype influences observability. It is natural to consider how mobile observers can exploit this dependence through maneuvers to maximize \ac{SSAT} performance. To that end, this work presents a trajectory planning method for low-thrust tracking platforms, aiming to optimize their joint navigation and tracking performance. It should be noted that maneuvers in themselves can cause uncertainties in the navigation solution; understanding the trade-off between the gain in navigation accuracy and the uncertainties induced by maneuver errors is an important consideration for future research. We assume, however,  that this effect is negligible in the present study. The main technical contributions of our work are described below.

We first develop an information-theoretic objective function that quantifies the expected performance of a sequential estimator that jointly determines the absolute states of an observer and target(s). The resulting quantity is functionally dependent on 1) the states of the constituent spacecraft, 2) the sensor modality, and 3) the system dynamics. Formally, this objective is defined as the mutual information between the augmented \ac{SSAT} system state and measurement, collected over an observation arc. For conciseness, we sometimes refer to this objective as the information gain. 

Our information-theoretic objective function is incorporated into a trajectory optimization problem that balances information gain and control effort. The balance is formed as a weighted average modulated by a single scalar homotopy factor. The problem is transformed into a static parameter optimization problem over the control inputs with the \ac{SCvx} \cite{Mao_2017} algorithm. Approximations of the mutual information based on Gaussian first- and second-order Taylor expansions of the augmented covariance are derived that are compatible with the problem at hand. Practical considerations for the transformation, such as proper regularization of discrete time nodes using the generalized Sundman transformation \cite{Szebehely_1969}, are also examined. 

We conclude by providing an analysis of our methods for cislunar \ac{SSAT} trajectory optimization. Without loss of generality, the motion of \acp{SO} in the cislunar region is modeled using the \ac{CRTBP}. The reader will note that the methods presented are broadly formulated to consider multiple observers and targets, with various sensing modalities; however, to highlight the key technical contributions presented here, we consider a test case with a single observer and target in neighboring \acp{DRO}. Other cases that demonstrate these additional capabilities are presented in the corresponding author's dissertation \cite{Wolf_2025}, and we anticipate disseminating these to the broader community in a future letter.  Collectively, these contributions support advancements in autonomous cislunar \ac{GNC} and surveillance. 

\subsection{Related Work}

The spaceflight community has devoted attention to problems similar to those in the present study. Previous work developed analytical approaches for determining optimal impulsive maneuvers applied to relative navigation using geometric definitions of observability \cite{Woffinden_2009, Grzymisch_2014, Franquiz_2018}. Related to our problem, Ref. \citenum{Greaves_2023} investigates impulsive maneuver design for optical \ac{SSAT} in cislunar space. The strategies mentioned above, however, do not consider the sensitivity of an estimator's performance with respect to these control inputs over extended time intervals (e.g., days or weeks), which may be necessary for some missions in cislunar space, nor do they consider low-thrust maneuvers. Alternatively, Ref. \cite{Pi_2014} incorporates relative navigation accuracy in extended trajectory design by enforcing observability constraints at discrete time nodes along a collocation grid and then minimizing control effort.

Information-theoretic definitions of optimality possess key benefits for the present problem. Important for our work, they are amenable to formally quantifying \ac{SSAT} accuracy over an extended trajectory arc so that the measure's quantity is functionally dependent on the control inputs. A large body of research has investigated the applicability of these measures in optimal experiment design, sensor tasking, and path planning \cite{Bell_1993, Emery_1998, Paninski_2005, Hero_2008, Bai_2021}. The Fisher information is particularly prevalent in the path planning literature \cite{Hammel_1989, Oshman_1999, Hou_2021}. It is easy to compute, and has the intuitive definition as the inverse \ac{CRLB}. However, as noted in Ref. \cite{Adurthi_2020}, the Fisher information quantifies the local curvature of the relative entropy, or \ac{KLD}, contained at the parameter estimate. Mutual information, however, is the \textit{expected value} of the \ac{KLD} between a prior and posterior distribution, and therefore, provides a global representation of the information gain. The distinction is important as the latter can better capture higher-order information resulting from nonlinear dynamics and measurements -- a critical component in quantifying expected \ac{SSAT} performance.

The present study solves the optimal \ac{SSAT} trajectory design problem via \ac{SCP}; specifically, using the \ac{SCvx} algorithm \cite{Mao_2018, Malyuta_2022}. State-of-the-art SCP algorithms have gained popularity in aerospace applications recently because they are generally robust to a coarse initial solution guess, require few solver iterations, and are amenable to off-the-shelf convex program algorithms that possess strong theoretical guarantees. In particular, they have enjoyed considerable success in real-time applications, including powered descent guidance \cite{Blackmore_2010, Szmuk_2017, Reynolds_2020}, aerocapture guidance \cite{Han_2019, Rataczak_2025}, model predictive control for spacecraft swarms \cite{Morgan_2013}, and autonomous uncertainty-aware spacecraft trajectory design \cite{Oguri_2024, Kumagai_2024}. 

\subsection{Organization}

The rest of this paper is organized as follows: The next section provides a mathematical description of the \ac{SSAT} problem and introduces the optimization framework used for trajectory generation in this study. Section \ref{sec:information_gain} proceeds to derive the information-based objective used in quantifying expected \ac{SSAT} performance. This section also provides approximations of this objective based on first- and second-order Taylor polynomial expansions of the system state error covariance matrix. Section \ref{sec:scvx} describes the \ac{SCvx} algorithm in the context of \ac{SSAT} trajectory optimization. Section \ref{sec:results} presents a numerical case study to demonstrate the applicability of our proposed approach to cislunar autonomous navigation and tracking operations. Finally, section \ref{sec:concluisons} concludes this work by summarizing our main contributions, the impact of these contributions, and areas for planned future research.  

%
\section{Problem Formulation}\label{sec:problem_formulation}
\subsection{Cislunar \acf{SSAT}}

\begin{figure}[hbt!]
    \centering
    \includegraphics[width=0.65\textwidth]{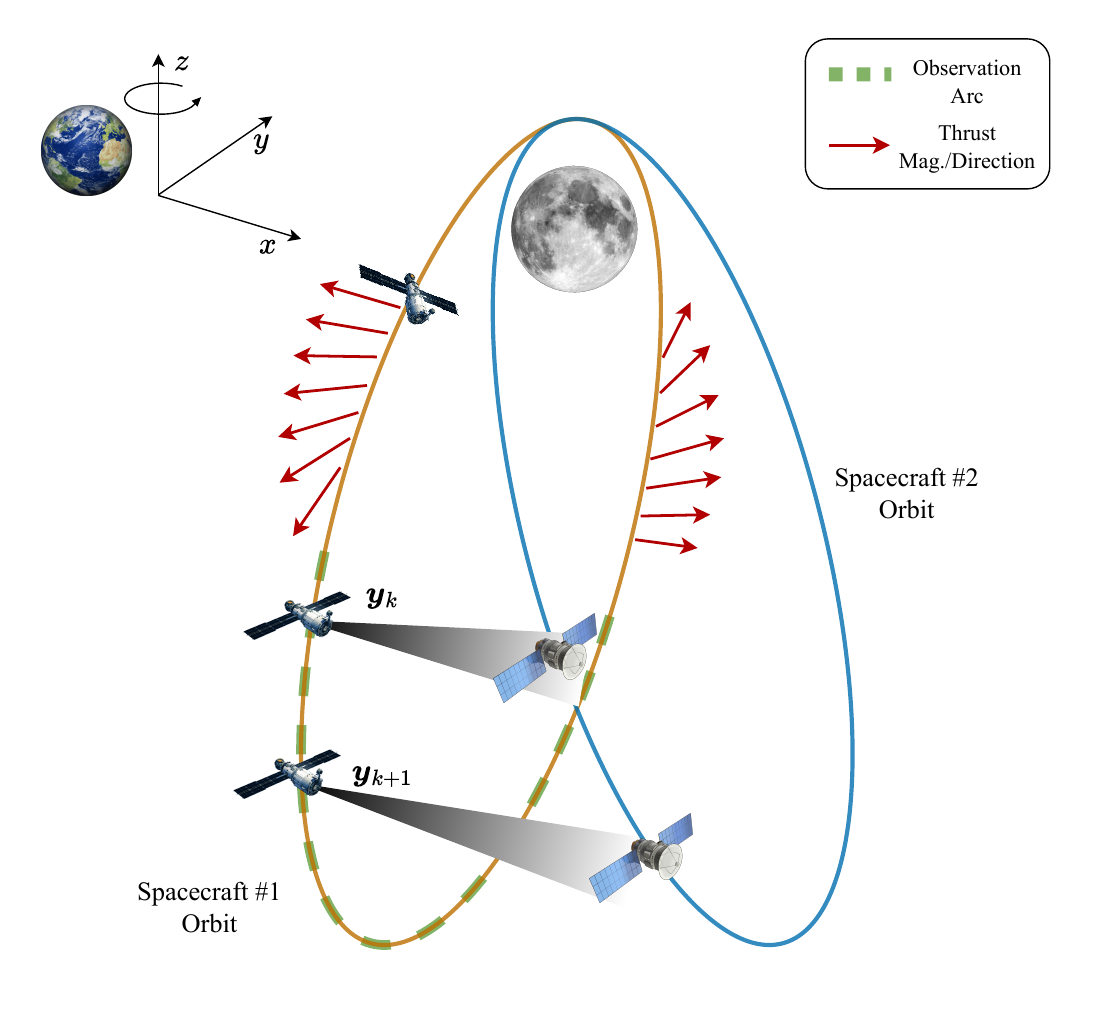}
    \caption{Illustration of the \acs{SSAT} problem for two spacecraft operating in proximal \acs{NRHO}s. The green dashed line is the observation arc. }\label{fig:SSAT_illustration}
\end{figure}

We formulate the \ac{SSAT} problem setup in a manner that is generalizable to multiple dynamical models and measurement modalities. It should be understood that this formulation applies to spacecraft operating in the cislunar domain, outfitted with various sensors (e.g., optical, radiometric, LIDAR). In Section \ref{sec:results}, we discuss the specific dynamical and measurement models used in our analysis. As an example to acclimate the reader, Figure \ref{fig:SSAT_illustration} illustrates the optical \ac{SSAT} problem for two spacecraft operating along paths that are proximal to a periodic orbit defined in the \ac{CRTBP}. In the current setup, one observer collects relative measurements of the other over a specified observation arc, indicated by the green dashed line. During the interim, the active spacecraft maneuvers to produce a favorable natural trajectory for the observation period. Illustrated here, only the observer can maneuver; however, this is not required for our formulation.


\subsubsection{Discrete Time Dynamics}
\ac{SSAT} estimates the absolute states of two or more spacecraft with relative measurements. We consider a sequential version of the problem so that the parameter of interest is an augmented state vector 
\begin{equation}
    \tilde{\boldsymbol{x}}_{k} = \left[ \boldsymbol{x}_{k}^{1 \top}, \boldsymbol{x}_{k}^{2 \top}, \cdots,  \boldsymbol{x}_{k}^{N_{s} \top} \right]^\top
\end{equation}
that concatenates the state of all spacecraft at discrete time $t_k$ contained in the set $\{\boldsymbol{x}^i_k\}_{i \in [N_s]}$. In the above, $\boldsymbol{x}_{k}^i \in \mathbb{R}^6$ is the dynamical state of the $i^{\text{th}}$ spacecraft. The integer $N_{s}$ is the number of spacecraft and the index set $[N_s] \coloneqq \{i \in \mathbb{Z}: 1 \leq i \leq N_s\}$. 

The translational motion of the augmented state vector flows according to the natural dynamics in the Earth-Moon three-body system between discrete times that fall within an observation arc. This can be written as
\begin{equation}
    \tilde{\boldsymbol{x}}_{k + 1} = \tilde{\boldsymbol{f}}_{k}(\tilde{\boldsymbol{x}}_{k}) + \tilde{\boldsymbol{\nu}}_{k}. \label{eqn:dynamics_evolution}
\end{equation}
In other words, the mobile spacecraft are constrained to be passive during an observation arc. We note here for clarity that the key problem that this study addresses is how to optimally maneuver the spacecraft(s) \textit{between} observation arcs to maximize \ac{SSAT} performance. The augmented discrete-time translational dynamics are generated by concatenating those for each spacecraft, 
\begin{equation}
    \tilde{\boldsymbol{f}}_{k}(\cdot) = [\boldsymbol{f}_{k}^{1 \top}(\cdot), \cdots, \boldsymbol{f}_{k}^{N_s \top}(\cdot)]^\top,
\end{equation}
where $\boldsymbol{f}_k^i(\cdot)$ are the integrated continuous time equations of motion between times $t_k$ and $t_{k + 1}$ for the $i^{\text{th}}$ spacecraft, i.e., 
\begin{equation}
    \boldsymbol{f}^i_k(\boldsymbol{x}_k^i) = \int_{t_k}^{t_{k + 1}}\boldsymbol{f}(\boldsymbol{x}^i(\tau), \tau)d\tau.
\end{equation}
The variable $\tilde{\boldsymbol{\nu}}_{k}$ is the augmented random process noise vector that accounts for unmodeled accelerations evolving between times $t_k$ and $t_{k + 1}$. It is distributed according to a zero-mean Gaussian distribution

\begin{equation}
    \tilde{\boldsymbol{\nu}}_{k} \sim \mathlarger{\mathcal{N}}\left( \boldsymbol{0}, \tilde{Q}_{k} \right).
\end{equation}
The augmented process noise covariance matrix is the block diagonal matrix 
\begin{equation}
    \tilde{Q}_{k} = \text{blkdiag}\left(\left[ Q_{k}^1, \cdots, Q_{k}^{N_{s}}  \right]\right), 
\end{equation}
where, 
\begin{equation}
    Q_{k}^i = \int_{t_k}^{t_{k + 1}} \Phi^i(t_{k + 1}, \tau) \Gamma Q \Gamma^\top \Phi^{i \top}(t_{k + 1}, \tau) d\tau. 
\end{equation}
The matrix $\Phi^{i}(\cdot)$ is the state transition matrix for the $i^{\text{th}}$ spacecraft that evolves according to the matrix differential equation 
\begin{equation}
    \dot{\Phi}^i(t, t_k) = \frac{\partial \boldsymbol{f}(\boldsymbol{x}^{i}(t), t)}{\partial \boldsymbol{x}^{i}(t)} \Phi^i(t, t_k),
\end{equation}
where $\boldsymbol{f}(\cdot)$ are the continuous time equations of motion, and $\Gamma$ is the process noise gain matrix. The matrix $Q$ is the unmodeled acceleration process noise \ac{PSD}. 


\subsubsection{Measurements}
One spacecraft collects relative measurements of the others. This is not a requirement, as all active spacecraft can possess sensors. Depending on the spacecraft configuration, additional geometric information can be provided with multiple sensor platforms \cite{Kruger_2024}, however, we omit this consideration for our study.  Without loss of generality, assume that the observer is indexed by the first spacecraft in the set $\{\boldsymbol{x}_k^i\}_{i \in [N_s]}$. Under the assumption that the single-target measurements, $\boldsymbol{y}_k^i \in \mathbb{R}^m$, are collected simultaneously for all target spacecraft, the augmented measurement vector is 
\begin{equation}
    \tilde{\boldsymbol{y}}_{k} = \left[\boldsymbol{y}^{2 \top}_{k}, \boldsymbol{y}^{3 \top}_{k}, \cdots, \boldsymbol{y}^{N_s \top}_{k}
    \right]^\top,
\end{equation}
where,
\begin{equation}
    \boldsymbol{y}_{k}^{i} = \boldsymbol{h}(\boldsymbol{x}_{k}^{i}, \boldsymbol{x}_{k}^1) + \boldsymbol{w}^i_k\mbox{,}\hspace{0.25cm} \text{for } i \in [N_S] \textbackslash\{1\}.
\end{equation}
Like the discrete-time dynamics, the augmented measurement function is expressed in short form as 
\begin{equation}
    \tilde{\boldsymbol{y}}_{k} = \tilde{\boldsymbol{h}}_{k}(\tilde{\boldsymbol{x}}_{k}) + \tilde{\boldsymbol{w}}_{k},
\end{equation}
where $\tilde{\boldsymbol{h}}_{k}(\cdot)$ stacks the single target measurement functions above. We assume that measurements of each target spacecraft are uncorrelated, so $\tilde{\boldsymbol{w}}_k$ is distributed according to 
\begin{equation}
    \tilde{\boldsymbol{w}_k} \sim \mathcal{N}(\boldsymbol{0}, \tilde{R}_k),
\end{equation}
where 
\begin{equation}
    \tilde{R}_k = \text{blkdiag}([\underbrace{R, \ldots, R}_{N_s - 1\ \text{times}}]).
\end{equation}
The matrix $R$ is the single target measurement noise covariance matrix. 


\subsection{Multi-Objective Optimal Control Problem}

This subsection provides a mathematical description of the optimal control problem. The approach balances information collection and control effort by forming an objective that is a weighted average of these two competing interests. This combination is controlled via a single scalar tuning parameter. For convenience and clarity, here the problem is expressed in continuous time. Later, we show how the optimal control problem is transformed into a static optimization problem using \ac{SCvx}.

The problem is a fixed-time two-point boundary-value optimal control problem of the form 
\begin{subequations}
    \begin{flalign}
        &\hspace{35ex} \min_{\{\boldsymbol{u}^j(t)\}_{j \in \mathcal{J}_c}} J \left( \{\boldsymbol{x}^j(t), \boldsymbol{u}^j(t)\}_{j\in \mathcal{J}_c}\right)&\label{eqn:original_cost}\\
        &\hspace{36ex}\mbox{s.t.} \hspace{1ex} \dot{\boldsymbol{x}}^{j}(t) = \boldsymbol{f}(\boldsymbol{x}^{j}(t), \boldsymbol{u}^{j}(t), t),&\\
        &\hspace{40ex} \boldsymbol{g}_{\mathrm{ic}}(\boldsymbol{x}^{j}(t_0)) = \boldsymbol{0}\mbox{,}&\\
        &\hspace{40ex} \boldsymbol{g}_{\mathrm{tc}}(\boldsymbol{x}^{j}(t_f)) = \boldsymbol{0}\mbox{,}&\\
        &\hspace{40ex} \|\boldsymbol{u}^{j}(t)\|_2 \leq a_{\text{max}}^{j}(t)\mbox{,}\hspace{0.25cm} \text{for } j \in \mathcal{J}_c\mbox{.}\label{eqn:nonlinear_thrust_constraint}&
    \end{flalign}
    \label{eqn:general_noncovex_continuous_problem}%
\end{subequations}
The integer set $\mathcal{J}_c \subseteq [N_s]$ indexes the spacecraft that are controllable. In other words, we assume that there is a subset of spacecraft in the total formation capable of maneuvering, i.e., $\{\boldsymbol{x}^j(t) \}_{j \in \mathcal{J}_c} \subseteq \{\boldsymbol{x}^i(t) \}_{i \in [N_s]}$ with corresponding controls $\{\boldsymbol{u}^j(t) \}_{j \in \mathcal{J}_c}$. Initial and terminal boundary constraints are defined as
\begin{subequations}
    \begin{flalign}
        \boldsymbol{g}_{\mathrm{ic}}(\boldsymbol{x}^{j}(t_0)) &= \boldsymbol{x}^{j}(t_0) - \boldsymbol{x}^{*j}_{0}\mbox{,}\\
        \boldsymbol{g}_{\mathrm{tc}}(\boldsymbol{x}^{j}(t_f)) &= \boldsymbol{x}^{j}(t_f) - \boldsymbol{x}^{* j}_{f},\hspace{0.25cm} \text{for } j \in \mathcal{J}_c,
    \end{flalign}
    \label{eqn:boundary_conditions}%
\end{subequations}
respectively, where $\boldsymbol{x}^{* j}_{0} \in \mathbb{R}^6$ and $\boldsymbol{x}^{* j}_{f} \in \mathbb{R}^6$ are predetermined initial and terminal states for the $j^{\text{th}}$ controllable spacecraft. As mentioned previously, the spacecraft are constrained to be passive during an observation arc. Therefore, a thrust acceleration condition is defined as 
\begin{equation}
    a_{\text{max}}^{j}(t) = \begin{cases}
        a_{\text{max}}^{j}, & \text{if } t \notin \mathcal{T}_{\text{obs}},\\
        0, & \text{otherwise.}
    \end{cases}
\end{equation}
The scalar variable $a_{\text{max}}^j$ is the maximum deliverable thrust acceleration for the $j^{\text{th}}$ spacecraft, and $\mathcal{T}_{\text{obs}}$ denotes the set of predetermined times corresponding to observation arcs. 

The objective in Eq. \ref{eqn:general_noncovex_continuous_problem} is expressed as 
\begin{equation}
    J \left( \{\boldsymbol{x}^j(t), \boldsymbol{u}^j(t)\}_{j\in \mathcal{J}_c}\right) = 
    \mathlarger{\int}_{0}^{t_f} \left[(1 - \alpha_h)\sum_{j \in \mathcal{J}_c}
    \|\boldsymbol{u}^{j}(t)\|_2 + \frac{\alpha_h}{\lambda_h} W\left( \{\boldsymbol{x}^j(t)\}_{j \in \mathcal{J_c}}; \{\boldsymbol{x}^\ell(t)\}_{\ell \in [N_s]\textbackslash \mathcal{J}_c} \right)\right] dt.
    \label{eqn:objective_general_form}
\end{equation}
which balances the total control input exerted by all spacecraft, and an information gain functional. For now, we express the latter with the placeholder function $W\left(\cdot \right)$ that is functionally dependent on the states of the active spacecraft, $\{\boldsymbol{x}^j(t)\}_{j \in \mathcal{J_c}}$, and parameterized by the inactive spacecraft, $\{\boldsymbol{x}^\ell(t)\}_{\ell \in [N_s]\textbackslash \mathcal{J}_c}$. The convex combination in the integrand is moderated through the homotopy parameter $\alpha_h \in [0, 1)$. The parameter $\lambda_h$ scales the information gain to be of the same order as the control effort. The next section provides a formal description of the information gain.

%
\section{Information-Based Objective Function}\label{sec:information_gain}


\subsection{Geometric/Dynamical Interpretation}

\begin{figure}[hbt!]
    \centering
    \includegraphics[width=0.95\textwidth]{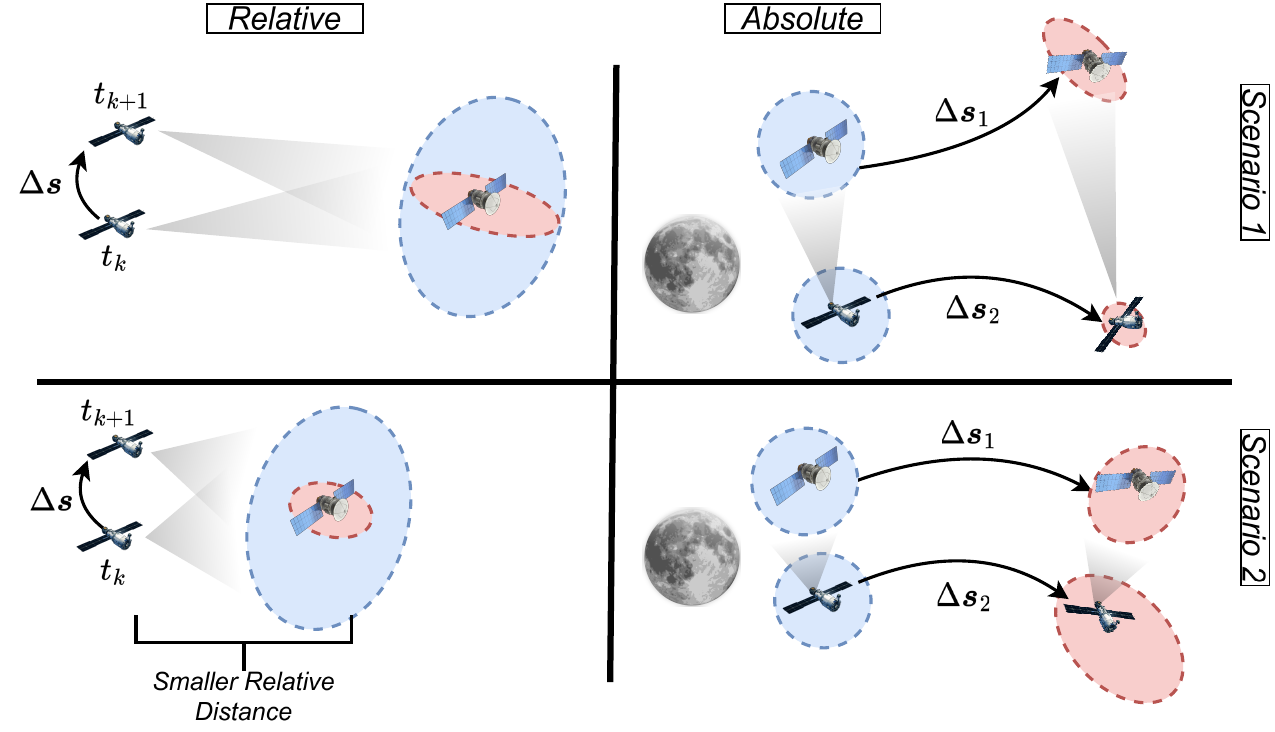}
    \caption{Illustration of geometric and dynamical differentiation for optical \ac{SSAT} observability with a relative (left) and absolute (right)  perspective. The blue and red ellipses represent the prior and posterior uncertainties. If the observer's state is known precisely, an information-optimal trajectory favors a closer approach, and the opposite when the observer's state is uncertain. }\label{fig:geometric_and_dynamical_interpretation}
\end{figure}

To provide context, a useful exercise is to understand the influence of information gain on \ac{SSAT} trajectory optimization through a geometric and dynamical perspective. This is most clearly apparent for an optical observer, and is illustrated in Figure \ref{fig:geometric_and_dynamical_interpretation}. The left-hand side illustrates a relative view of the problem where the target is fixed in the frame, and the right-hand side an absolute view. Two scenarios are presented: the top and bottom panels, which correspond to scenarios with larger and smaller baselines between the observer and target, respectively. The translational displacement between time steps is represented by the variable $\Delta\boldsymbol{s}$. Prior and posterior uncertainties are depicted with the blue and red ellipses, respectively. In the relative view, assuming perfect knowledge of the observer state, the greatest information gain is generated by operating the observer as close as possible to the target. This is intuitive, as decreasing the relative distance increases the geometric diversity in a sequence of relative measurements (i.e., for the same translational displacement, a larger change in the bearing angles). 

On the other hand, determining the absolute states of both the target and observer requires sufficient \textit{dynamical diversity} between the target and observer. In this sense, it pays to operate with a greater baseline distance. As will be shown, the mode of operation largely depends on the observer uncertainty so that 1) when the observer's state is well constrained, the information gain function prioritizes geometric diversity and incentivizes the observer to choose a path closer to the target, and 2) in lacking a well constrained navigation solution, the information gain function prioritizes dynamical diversity, directing the observer to operate with a greater baseline distance. 

\subsection{Introducing the Objective Function}

To ease description, for now, consider one augmented measurement collected at time step $t_k$. The mutual information for this single measurement trial is defined as
\begin{equation}
    I(\tilde{\boldsymbol{x}}_{k}; \tilde{\boldsymbol{y}}_{k}) = D_{\text{KL}}\left(p(\tilde{\boldsymbol{x}}_{k}, \tilde{\boldsymbol{y}}_{k})\|p(\tilde{\boldsymbol{x}}_{k})p(\tilde{\boldsymbol{y}}_{k})\right)\mbox{,}
        \label{eqn:MI_joint_KLD}
\end{equation}
where $D_{\text{KL}}(p(\cdot) \| q(\cdot))$ is the \acf{KLD},
\begin{equation}
    D_{\text{KL}}(p(\xi)||q(\xi)) =  \int_{-\infty}^\infty p(\xi) \ln\left(\frac{p(\xi)}{q(\xi)}\right) d \xi,
\end{equation}
for arbitrary continuous distributions $p(\cdot)$ and $q(\cdot)$. The \ac{KLD} is commonly interpreted as a measure of dissimilarity between probability distributions. Marginalizing the joint distribution in Eq.~\ref{eqn:MI_joint_KLD}, the mutual information can be written as
\begin{flalign}
    I(\tilde{\boldsymbol{x}}_{k}; \tilde{\boldsymbol{y}}_{k}) &= \mathbb{E}_{\tilde{\boldsymbol{y}}_{k}}\left[D_{\text{KL}}\left(p(\tilde{\boldsymbol{x}}_{k}|\tilde{\boldsymbol{y}}_{k})\|p(\tilde{\boldsymbol{x}}_{k})\right)\right] \nonumber\\
    &= \mathbb{E}_{\tilde{\boldsymbol{x}}_{k}}\left[D_{\text{KL}}\left(p(\tilde{\boldsymbol{y}}_{k}|\tilde{\boldsymbol{x}}_{k})\|p(\tilde{\boldsymbol{y}}_{k})\right)\right].\label{eqn:marginal_MI}
\end{flalign}
The operator $\mathbb{E}[\cdot]_\xi$ is the statistical expectation over the distribution $p(\xi)$. Inspecting Eq. \ref{eqn:marginal_MI}, we find that maximizing the mutual information amounts to maximizing the expected dissimilarity between a prior and its conditional distribution. The intuition for its use as an objective function in our context comes from the fact that the conditional density \textit{always} contains the same or greater information content of the prior, provided that the conditional expectation results in an unbiased estimator, as is assumed here. Therefore, optimally discriminating between the two provides the greatest information gain. 

\subsection{Mutual Information Under Gaussian First- and Second-Order Covariance Approximations}\label{sec:MI_first_order_approximation}

We employ two approximate forms of mutual information, based on first- and second-order Taylor expansions of the time-aggregate system state and measurement covariance matrices. To begin this part of the discussion, consider the framework presented previously for a single measurement trial applied to all measurements collected over an observation interval. The sequence of augmented states, $\{\tilde{\boldsymbol{x}}_k\}_{k = p}^{p + N_\text{meas}}$, and measurements, $\{\tilde{\boldsymbol{y}}_k\}_{k = p}^{p + N_{\text{meas}}}$, collected during this period, each stacked into a vector form, are
\begin{equation}
    \tilde{\boldsymbol{X}} = \left[\tilde{\boldsymbol{x}}_{p}^\top, \cdots, \tilde{\boldsymbol{x}}_{p + N_{\text{meas}}}^\top \right]^\top
\end{equation}
and
\begin{equation}
    \tilde{\boldsymbol{Y}} = \left[ \tilde{\boldsymbol{y}}_{p}^\top, \cdots, \tilde{\boldsymbol{y}}_{p + N_{\text{meas}}}^\top \right].
\end{equation}
The integer $p$ indexes the discrete time $t_p$ coinciding with the start of the measurement period, and $N_{\text{meas}}$ is the number of measurements. The present study assumes that $\tilde{\boldsymbol{X}}$ and $\tilde{\boldsymbol{Y}}$ are jointly Gaussian so that
\begin{equation}
    p(\tilde{\boldsymbol{X}}, \tilde{\boldsymbol{Y}}) = \mathlarger{\mathcal{N}}\left(\begin{bmatrix}
        \tilde{\boldsymbol{X}}\\
        \tilde{\boldsymbol{Y}}
    \end{bmatrix};
    \begin{bmatrix}
        \bar{\tilde{\boldsymbol{X}}}\\
        \bar{\tilde{\boldsymbol{Y}}}
    \end{bmatrix},
    \begin{bmatrix}
        \Sigma_{\tilde{\boldsymbol{X}}\tilde{\boldsymbol{X}}} & \Sigma_{\tilde{\boldsymbol{X}}\tilde{\boldsymbol{Y}}}\\
        \Sigma^\top_{\tilde{\boldsymbol{X}}\tilde{\boldsymbol{Y}}} & \Sigma_{\tilde{\boldsymbol{Y}}\tilde{\boldsymbol{Y}}}
    \end{bmatrix}
    \right),
\end{equation}
which, in short form, is equivalently expressed as 
\begin{equation}
    p(\tilde{\boldsymbol{Z}}) = \mathlarger{\mathcal{N}}(\tilde{\boldsymbol{Z}}; \bar{\tilde{\boldsymbol{Z}}}, \Sigma_{\tilde{\boldsymbol{Z}}\tilde{\boldsymbol{Z}}})\label{eqn:big_pz}.
\end{equation}
The mutual information between the stacked time-aggregate augmented state, $\tilde{\boldsymbol{X}}$, and measurement, $\tilde{\boldsymbol{Y}}$, becomes 
\begin{subequations}
    \begin{flalign}
        I(\tilde{\boldsymbol{X}}; \tilde{\boldsymbol{Y}}) &= 
        \frac{1}{2}\ln\left(\frac{|\Sigma_{\tilde{\boldsymbol{X}}\tilde{\boldsymbol{X}}}||\Sigma_{\tilde{\boldsymbol{Y}}\tilde{\boldsymbol{Y}}}|}{|\Sigma_{\tilde{\boldsymbol{Z}}\tilde{\boldsymbol{Z}}}|} \right)\label{eqn:MI_Gaussian_first}\\
        &= \frac{1}{2}\ln \left(\frac{|\Sigma_{\tilde{\boldsymbol{X}}\tilde{\boldsymbol{X}}}|}{|\Sigma_{\tilde{\boldsymbol{X}}\tilde{\boldsymbol{X}}} - \Sigma_{\tilde{\boldsymbol{X}}\tilde{\boldsymbol{Y}}} \Sigma^{-1}_{\tilde{\boldsymbol{Y}}\tilde{\boldsymbol{Y}}} \Sigma^\top_{\tilde{\boldsymbol{X}}\tilde{\boldsymbol{Y}}}|}\right)\label{eqn:MI_Gaussian_second}\\
        &= \frac{1}{2}\ln \left(\frac{|\Sigma_{\tilde{\boldsymbol{Y}}\tilde{\boldsymbol{Y}}}|}{|\Sigma_{\tilde{\boldsymbol{Y}}\tilde{\boldsymbol{Y}}} - \Sigma^\top_{\tilde{\boldsymbol{X}}\tilde{\boldsymbol{Y}}} \Sigma^{-1}_{\tilde{\boldsymbol{X}}\tilde{\boldsymbol{X}}} \Sigma_{\tilde{\boldsymbol{X}}\tilde{\boldsymbol{Y}}}|}\right).\label{eqn:MI_Gaussian_third}
    \end{flalign}\label{eqn:MI_Gaussian}
\end{subequations}

There exists a symmetry of the arguments in the second and third expressions. While their quantities are equivalent, the former describes the reduction in uncertainty of the state given the measurement, and the latter describes the reduction in uncertainty in the measurement given the state. For reasons that will become clear, the third expression, i.e., that in Eq. \ref{eqn:MI_Gaussian_third}, is better suited for the present study. 

The present task is to determine the covariance matrices contained in Eq. \ref{eqn:MI_Gaussian}. First, form an explicit expression for the vector $\tilde{\boldsymbol{Z}}$ through the recursive function $\boldsymbol{G}: \mathbb{R}^M \rightarrow \mathbb{R}^M$, where the integer $M = N_\text{meas}\times(6 + m)$, such that
\begin{equation}
    \underbrace{
    \begin{bmatrix}
        \tilde{\boldsymbol{x}}_{p} \\
        \tilde{\boldsymbol{x}}_{p + 1} \\
        \vdots\\
        \tilde{\boldsymbol{x}}_{p + N_{\text{meas}}} \\
        \tilde{\boldsymbol{y}}_{p} \\
        \tilde{\boldsymbol{y}}_{p + 1}\\
        \vdots\\
        \tilde{\boldsymbol{y}}_{p + N_{\text{meas}}}
    \end{bmatrix}}_{\mathlarger{\tilde{\boldsymbol{Z}}}} = 
    \underbrace{
    \begin{bmatrix}
        \boldsymbol{0}\\
        \tilde{\boldsymbol{f}}_{p}(\tilde{\boldsymbol{x}}_{p})\\
        \vdots \\
        \tilde{\boldsymbol{f}}_{p + N_{\text{meas}} - 1}(\tilde{\boldsymbol{x}}_{p + N_{\text{meas}} - 1})\\
        \tilde{\boldsymbol{h}}_{p}(\tilde{\boldsymbol{x}}_{p})\\
        \tilde{\boldsymbol{h}}_{p + 1}(\tilde{\boldsymbol{x}}_{p + 1})\\
        \vdots \\
        \tilde{\boldsymbol{h}}_{p + N_{\text{meas}}}(\tilde{\boldsymbol{x}}_{p + N_{\text{meas}}})
    \end{bmatrix} + 
    \overbrace{
    \begin{bmatrix}
        \tilde{\boldsymbol{x}}_{p}\\
        \tilde{\boldsymbol{\nu}}_{p}\\
        \vdots \\
        \tilde{\boldsymbol{\nu}}_{p + N_{\text{meas}} - 1}\\
        \tilde{\boldsymbol{w}}_{p}\\
        \tilde{\boldsymbol{w}}_{p + 1}\\
        \vdots\\
        \tilde{\boldsymbol{w}}_{p + N_{\text{meas}}}
    \end{bmatrix}.}^{\mathlarger{\tilde{\boldsymbol{\Xi}}}} }_{\mathlarger{\boldsymbol{G}(\tilde{\boldsymbol{\Xi}})}}
\end{equation}
The stacked vector $\tilde{\boldsymbol{\Xi}}$ concatenates the initial augmented state at the start of the measurement period, and the process noise and measurement noise vectors over the observation period. The joint covariance matrix can then be expressed as 
\begin{flalign}
    \Sigma_{\tilde{\boldsymbol{Z}} \tilde{\boldsymbol{Z}}} &= \mathbb{E}[(\tilde{\boldsymbol{Z}} - \bar{\tilde{\boldsymbol{Z}}})(\tilde{\boldsymbol{Z}} - \bar{\tilde{\boldsymbol{Z}}})^\top]\nonumber\\
    &= \mathbb{E}\left[\left(\boldsymbol{G}(\tilde{\boldsymbol{\Xi}}) - \mathbb{E}[\boldsymbol{G}(\tilde{\boldsymbol{\Xi}})] \right)\left(\boldsymbol{G}(\tilde{\boldsymbol{\Xi}}) - \mathbb{E}[\boldsymbol{G}(\tilde{\boldsymbol{\Xi}})] \right)^\top\right].\label{eqn:joint_covariance_ZZ}
\end{flalign}
To evaluate Eq. \ref{eqn:joint_covariance_ZZ}, we can approximate the function $\boldsymbol{G}(\cdot)$ with a second-order Taylor expansion so that
\begin{equation}
    \boldsymbol{G}(\tilde{\boldsymbol{\Xi}}) \simeq \boldsymbol{G}(\bar{\tilde{\boldsymbol{\Xi}}}) + \nabla \boldsymbol{G} \cdot \delta \tilde{\boldsymbol{\Xi}} + \boldsymbol{B}.
\end{equation}
In the above, $\bar{\tilde{\boldsymbol{\Xi}}} = \mathbb{E}[\tilde{\boldsymbol{\Xi}}]$ and $ \delta\tilde{\boldsymbol{\Xi}} = \tilde{\boldsymbol{\Xi}} - \bar{\tilde{\boldsymbol{\Xi}}}$. The vector $\boldsymbol{B}$ is determined by
\begin{equation}
    \boldsymbol{B} = \frac{1}{2} \begin{bmatrix}
        \text{tr}\left(\nabla^2\boldsymbol{G}_1 \cdot \delta \tilde{\boldsymbol{\Xi}} \delta \tilde{\boldsymbol{\Xi}}^\top \right)\\
        \text{tr}\left(\nabla^2\boldsymbol{G}_2\cdot \delta \tilde{\boldsymbol{\Xi}} \delta \tilde{\boldsymbol{\Xi}}^\top \right)\\
        \vdots\\
        \text{tr}\left(\nabla^2\boldsymbol{G}_M \cdot \delta \tilde{\boldsymbol{\Xi}} \delta \tilde{\boldsymbol{\Xi}}^\top \right)\\
    \end{bmatrix}
\end{equation}
The Jacobian matrix of $\boldsymbol{G}(\cdot)$ is
\begin{equation}
    \nabla\boldsymbol{G} = \frac{\partial \boldsymbol{G}(\tilde{\boldsymbol{\Xi}})}{\partial\tilde{\boldsymbol{\Xi}}} \bigg|_{\tilde{\boldsymbol{\Xi}} = \bar{\tilde{\boldsymbol{\Xi}}}},
\end{equation}
and the Hessian matrix of the $i^{\text{th}}$ vector component of $\boldsymbol{G}(\cdot)$ is
\begin{equation}
    \nabla^2\boldsymbol{G}_i = \frac{\partial^2 \boldsymbol{G}_i(\tilde{\boldsymbol{\Xi}})}{\partial \boldsymbol{G}_i(\tilde{\boldsymbol{\Xi}})\partial\boldsymbol{G}_i(\tilde{\boldsymbol{\Xi}})^\top} \bigg|_{\tilde{\boldsymbol{\Xi}} = \bar{\tilde{\boldsymbol{\Xi}}}}.
\end{equation}
Note that we drop the function argument in the derivatives for conciseness. 
\begin{lemma}
\text{(Ref. \citenum{Athans_1968}})  If $\boldsymbol{\xi} \in \mathbb{R}^n$ is a zero-mean Gaussian vector-valued random variable with covariance $\Sigma$, then for any two square matrices $C \in \mathbb{R}^{n \times n}$ and $D \in \mathbb{R}^{n \times n}$,
\begin{equation}
    \mathbb{E}\left[\boldsymbol{\xi}~\text{tr}\left(C \boldsymbol{\xi} \boldsymbol{\xi}^\top \right)\right] = \boldsymbol{0}
\end{equation}
and 
\begin{align}
    \mathbb{E}\left[ \text{tr}\left( 
    C \boldsymbol{\xi} \boldsymbol{\xi}^\top D \boldsymbol{\xi} \boldsymbol{\xi}^\top \right) \right] &= \mathbb{E}\left[\text{tr}\left(C \boldsymbol{\xi} \boldsymbol{\xi}^\top\right) \text{tr}\left(D \boldsymbol{\xi} \boldsymbol{\xi}^\top\right)\right] \nonumber \\
    &= 2~\text{tr}\left(C \Sigma D \Sigma \right) + \text{tr}\left( C \Sigma \right) \text{tr}\left( D \Sigma \right).
\end{align}
\end{lemma}

From the above lemma, it follows that the covariance can be approximated to second-order with
\begin{equation}
    \Sigma_{\tilde{\boldsymbol{Z}} \tilde{\boldsymbol{Z}}} \simeq \mathbb{E}\left[\nabla \boldsymbol{G}\delta \tilde{\boldsymbol{\Xi}} \delta \tilde{\boldsymbol{\Xi}}^\top\nabla \boldsymbol{G}^\top + \boldsymbol{B}\boldsymbol{B}^{\top}  \right], \label{eqn:approximate_Sigma_ZZ}
\end{equation}
and taking the expectation, 
\begin{equation}
    \mathlarger{\Sigma}_{\tilde{\boldsymbol{Z}} \tilde{\boldsymbol{Z}}} \simeq \underbrace{ \nabla\boldsymbol{G} \mathlarger{\Sigma}_{ \tilde{\boldsymbol{\Xi}} \tilde{\boldsymbol{\Xi}}} \nabla \boldsymbol{G}^\top}_{\text{First-Order Approximation}} + \mathcal{B}.\label{eqn:second_order_MI_approx}
\end{equation}
Each element of the matrix $\mathlarger{\mathcal{B}} \in \mathbb{R}^{M\times M}$ is determined by
\begin{equation}
    \mathcal{B}_{i, j} = 2~\text{tr}\left( \nabla^2\boldsymbol{G}_i \Sigma_{\tilde{\boldsymbol{\Xi}} \tilde{\boldsymbol{\Xi}}} \nabla^2\boldsymbol{G}_j \Sigma_{\tilde{\boldsymbol{\Xi}} \tilde{\boldsymbol{\Xi}}} \right) + \\
    \text{tr}\left( \nabla^2\boldsymbol{G}_i \Sigma_{ \tilde{\boldsymbol{\Xi}} \tilde{\boldsymbol{\Xi}}} \right)\text{tr}\left( \nabla^2\boldsymbol{G}_j \Sigma_{ \tilde{\boldsymbol{\Xi}} \tilde{\boldsymbol{\Xi}}} \right), 
\end{equation}
and the covariance of $\tilde{\boldsymbol{\Xi}}$ is
\begin{flalign}
    \Sigma_{\tilde{\boldsymbol{\Xi}}\tilde{\boldsymbol{\Xi}}} &= \mathbb{E}\left[\left(\tilde{\boldsymbol{\Xi}} - \bar{\tilde{\boldsymbol{\Xi}}} \right)\left(\tilde{\boldsymbol{\Xi}} - \bar{\tilde{\boldsymbol{\Xi}}} \right)^\top\right]\nonumber\\
    &=\text{blkdiag}\left([\tilde{P}_1, \tilde{Q}_1, \cdots, \tilde{Q}_{N_{\text{meas}} - 1}, \tilde{R}_1, \cdots \tilde{R}_{N_\text{meas}}] \right).
\end{flalign}

The first term in Eq. \ref{eqn:second_order_MI_approx} provides a first-order approximation of the covariance, and including the matrix $\mathcal{B}$ produces a second-order approximation. In practice, it is easy to determine the derivatives $\nabla\boldsymbol{G}$ and $\nabla^2 \boldsymbol{G}_i$ with automatic differentiation. 


\subsection{Comparison of First- and Second-Order Moment Approximations}

\begin{figure}[hbt!]
    \centering
    \includegraphics[width=0.5\textwidth]{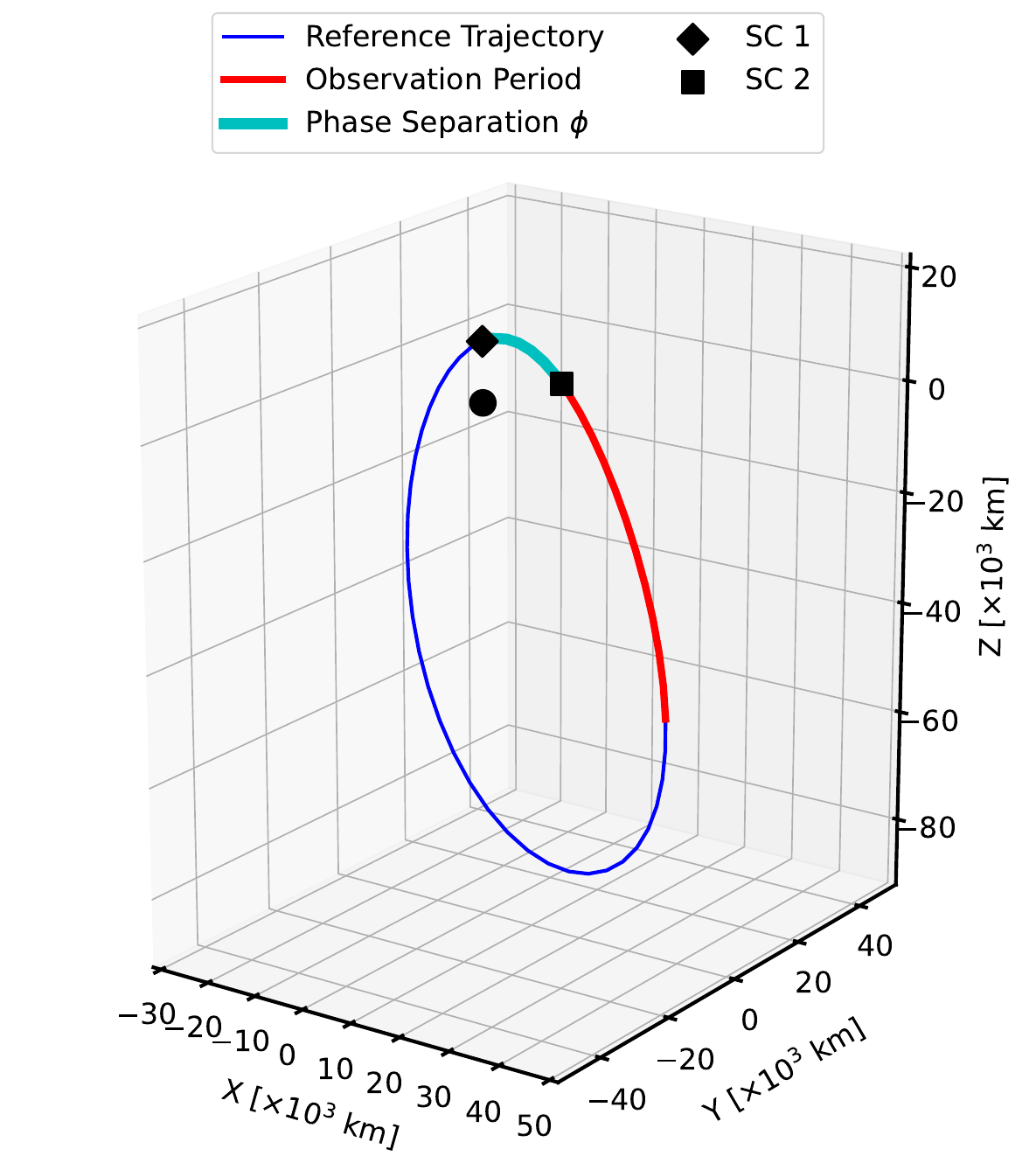}
    \caption{Illustration of the current experiment setup. Two spacecraft (SC~1 and SC~2) are separated by a phase angle $\phi$. The observer collects measurements of the target over a predetermined arc, indicated by the red line. }\label{fig:MI_experiment}
\end{figure}

%
\begin{figure}[hbt!]
    \centering
    \includegraphics[width=0.75\textwidth]{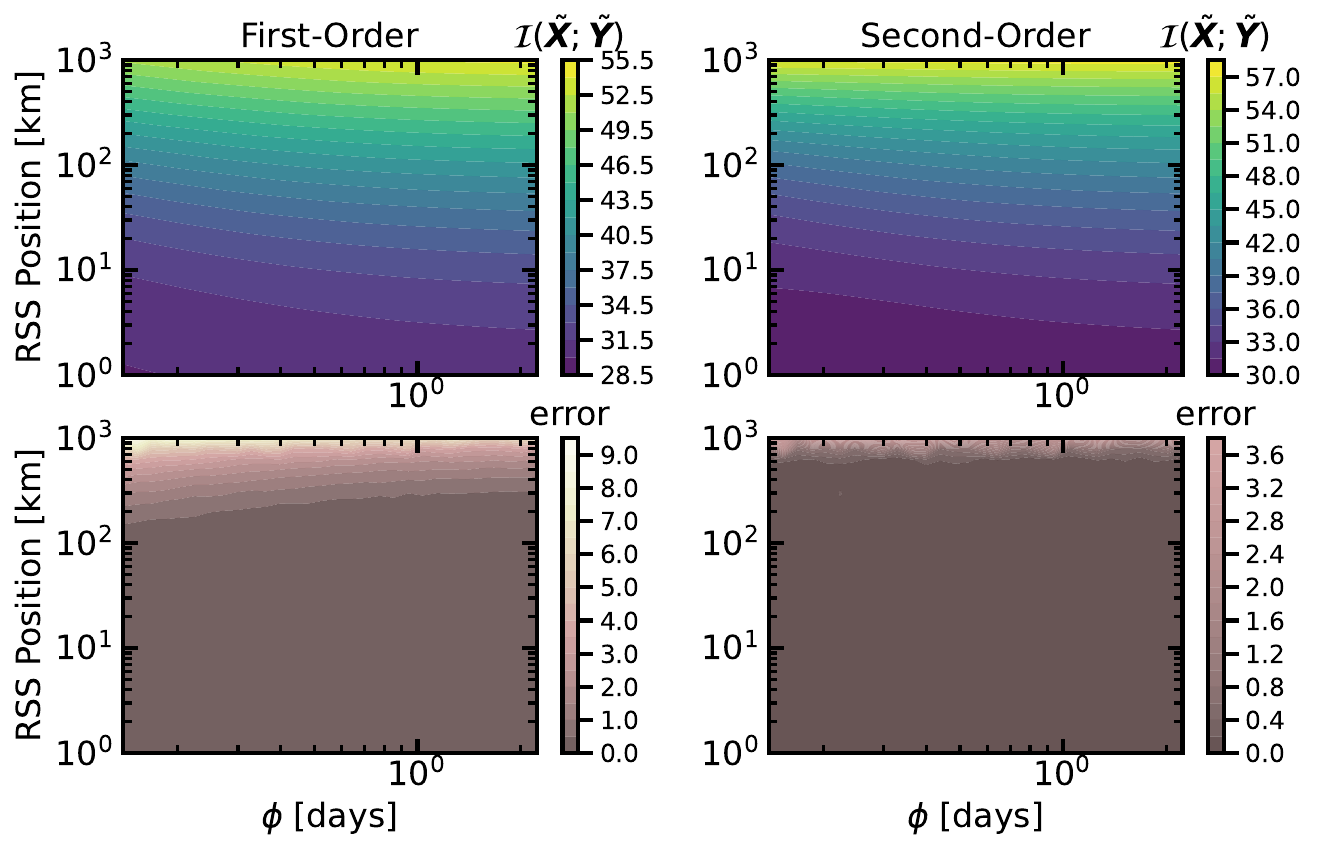}
    \caption{Mutual information for relative position measurements generated using a first-order covariance approximation (left), and a second-order covariance approximation (right) as a function of phase angle and prior observer uncertainty. The lower panels are the absolute errors of each approximation with respect to a truth estimate generated through Monte Carlo sampling. }\label{fig:MI_comparison_rel_pos}
\end{figure}
%
\begin{figure}[hbt!]
    \centering
    \includegraphics[width=0.75\textwidth]{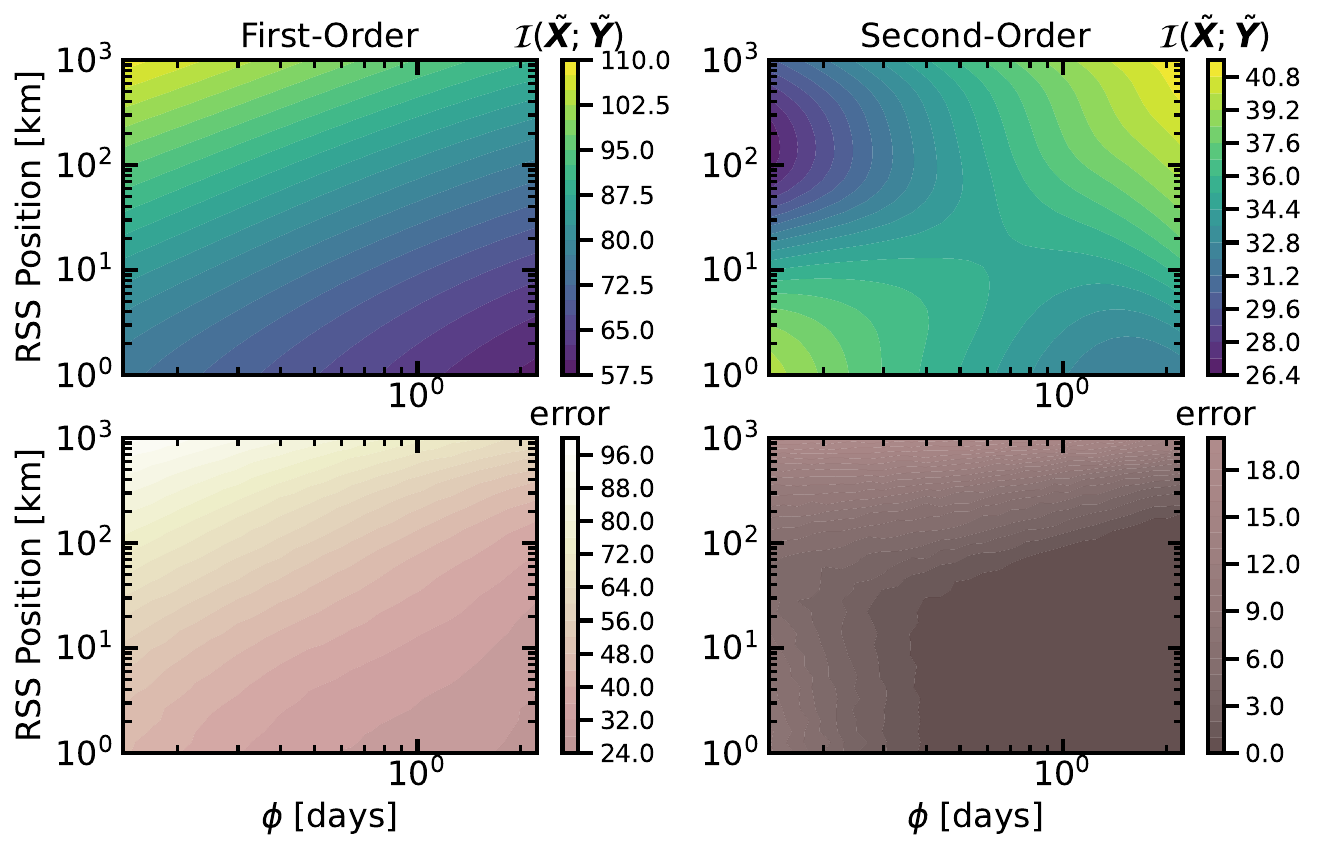}
    \caption{Mutual information for an optical observer generated using a first-order covariance approximation (left), and a second-order covariance approximation (right) as a function of phase angle and prior observer uncertainty. The lower panels are the absolute errors of each approximation with respect to a truth estimate generated through Monte Carlo sampling.}\label{fig:MI_comparison_optical}
\end{figure}

This discussion provides a comparison of the first- and second-order covariance approximations derived above for computing the mutual information. Figure \ref{fig:MI_experiment} illustrates the experiment setup. Two spacecraft are separated by a phase angle in a L2 southern \ac{NRHO}. Over the red portion of the reference orbit, the observer, indicated by ``SC~1", collects measurements of the target, indicated by ``SC~2". The objective of this experiment is to assess the relation between information gain and the relative geometry, as well as the initial state uncertainties associated with the spacecraft. It is unreasonable to evaluate all geometric and uncertainty configurations for this analysis.  However, to build intuition for how these factors affect the information gain, we alter the initial phase separation between the spacecraft and the initial uncertainty of the observer spacecraft.  

Two measurement modes are considered: relative position and optical bearing angles/angle rates. Figure \ref{fig:MI_comparison_rel_pos} plots the mutual information over the phase/uncertainty space generated with relative position measurements.  We notice that a larger initial uncertainty results in greater information gain. Also, increasing the relative distance, or phase angle in this example, always correlates with a greater information gain, regardless of the observer uncertainty. The latter can be attributed to the fact that relative positions are a linear function of the system state, so there is no geometrical advantage in decreasing the relative distance. The bottom panels show the absolute error of the first- and second-order approximations with respect to a truth estimate. We produce the truth estimate by evaluating  Eq. \ref{eqn:MI_Gaussian_third} with an empirical covariance using $10^4$ Monte Carlo samples given as
\begin{equation}
    \Sigma_{\tilde{\boldsymbol{Z}}\tilde{\boldsymbol{Z}}}^{\text{MC}} = \frac{1}{N_{\text{samp}} - 1}\sum_{i = 1}^{N_{\text{samp}}} \left(\boldsymbol{G}(\tilde{\boldsymbol{\Xi}}^i) - \bar{\boldsymbol{G}}^{\text{MC}}\right)\left(\boldsymbol{G}(\tilde{\boldsymbol{\Xi}}^i) - \bar{\boldsymbol{G}}^{\text{MC}}\right)^\top,
\end{equation}
where the mean vector is
\begin{equation}
    \bar{\boldsymbol{G}}^{\text{MC}} = \frac{1}{N_{\text{samp}}}\sum_{i = 1}^{N_\text{samp}} \boldsymbol{G}(\tilde{\boldsymbol{\Xi}}^i).
\end{equation}
Both the first- and second-order approximations are fairly consistent over the parameter space. At large initial uncertainties, the first-order approximation deviates slightly more from the truth estimate. 

Figure \ref{fig:MI_comparison_optical} displays the same results, but for an optical observer. Unlike previously, there are significant differences between the first- and second-order approximations in this example. In cases where the observer's state is well-known, the first-order approximation is adequate in approximating the information gain, and we see a smaller and fairly uniform trend in absolute error with respect to phase angle in this regime. However, with a larger initial uncertainty, the first-order approximation fails to produce a mutual information estimate consistent with the problem of \textit{simultaneously} estimating the state of the observer and target. We can infer that for optical \ac{SSAT} it is 1) best to operate in proximity when the observer's state is well constrained to maximize information gain, and 2) in the absence of a well-constrained state estimate, the observer should operate at a larger baseline distance with respect to the target. This result aligned with our previous expectation. 

This experiment provides a coarse analysis of the role of information gain within the \ac{SSAT} problem. In the subsequent sections, we discuss the specifics of how this objective is incorporated into an optimal control problem.

%
\section{\acf{SCvx} for Information-Optimal Cislunar Trajectory Generation}\label{sec:scvx}
We describe an approach for transforming the continuous time optimal control problem presented in Eq.~\ref{eqn:general_noncovex_continuous_problem}  into a sequence of convex parameter optimization problems using the \ac{SCvx} algorithm: an \ac{SCP} approach with global convergence and superlinear convergence rate guarantees \cite{Mao_2018}.

\subsection{Linearization}
Consider a linearized, continuous-time approximation of the optimal control problem in Eq. \ref{eqn:general_noncovex_continuous_problem}:
\begin{subequations}
    \begin{flalign}
        &\hspace{25ex} \min_{\{\boldsymbol{u}^j(t)\}_{j \in \mathcal{J}_c}} J \left(\{\boldsymbol{x}^j(t), \boldsymbol{u}^j(t)\}_{j \in \mathcal{J}_c} \right)&\\
        &\hspace{26ex} \mbox{s.t.} \hspace{1ex} \dot{\boldsymbol{x}}^j(t) = A^j(t)\boldsymbol{x}^j(t) + B^j(t)\boldsymbol{u}^j(t) + \boldsymbol{r}^j(t) \mbox{,}&\\
        &\hspace{30ex} \boldsymbol{g}_{\text{ic}}(\boldsymbol{x}^j(t_0)) = \boldsymbol{0}\mbox{,}&\\
        &\hspace{30ex} \boldsymbol{g}_{\text{tc}}(\boldsymbol{x}^j(t_f)) = \boldsymbol{0}\mbox{,}&\\
        &\hspace{30ex} \|\boldsymbol{u}^j(t)\|_2 \leq a^j_{\text{max}}(t)\mbox{,} &\\
        &\hspace{30ex} \|\delta \boldsymbol{x}^j(t)\|_2 + \|\delta \boldsymbol{u}^j(t)\|_2 \leq \eta, \hspace{2ex} \text{for } j \in \mathcal{J}_c. \label{eqn:continous_time_step_contraint}&
    \end{flalign}
    \label{eqn:linearized_continuous_form}%
\end{subequations}
Let the continuous time reference state and control variables for the $j^{\text{th}}$ controllable spacecraft be defined as $\bar{\boldsymbol{x}}^j(t)$ and $\bar{\boldsymbol{u}}^j(t)$, then 
\begin{subequations}
    \begin{flalign}
        A^j(t) &\coloneqq \frac{\partial \boldsymbol{f}(\boldsymbol{x}^j(t), \boldsymbol{u}^j(t), t)}{\partial \boldsymbol{x}^j(t)} \bigg|_{\bar{\boldsymbol{x}}^j(t), \bar{\boldsymbol{u}}^j(t)}\mbox{,}\\
        B^j(t) &\coloneqq \frac{\partial \boldsymbol{f}(\boldsymbol{x}^j(t), \boldsymbol{u}^j(t), t)}{\partial \boldsymbol{u}^j(t)}\bigg|_{\bar{\boldsymbol{x}}^j(t), \bar{\boldsymbol{u}}^j(t)} \mbox{,}\\
        \boldsymbol{r}^j(t) &\coloneqq \boldsymbol{f}(\bar{\boldsymbol{x}}^j(t), \bar{\boldsymbol{u}}^j(t), t) - A^j(t)\bar{\boldsymbol{x}}^j(t) - B^j(t)\bar{\boldsymbol{u}}^j(t)\mbox{.}
    \end{flalign}
    \label{eqn:linearization_definitions}%
\end{subequations}
In the above, $\boldsymbol{f}(\cdot)$ are the continuous-time dynamics of the three-body system, accounting for thrust acceleration. Linearization accuracy is only valid in a local neighborhood of the reference state and control variables for each spacecraft. Therefore, \ac{SCvx} imposes a trust region constraint  with Eq.~\ref{eqn:continous_time_step_contraint}, where 
\begin{subequations}
    \begin{flalign}
        \delta \boldsymbol{x}^j(t) &= \boldsymbol{x}^j(t) - \bar{\boldsymbol{x}}^j(t)\mbox{,}\\
        \delta \boldsymbol{u}^j(t) &= \boldsymbol{u}^j(t) - \bar{\boldsymbol{u}}^j(t)\mbox{.}
    \end{flalign}
    \label{eqn:delta_states_and controls}%
\end{subequations}
The scalar trust region radius $\eta$ is recursively adjusted with each convex program iteration, and the adjustment procedure is provided Ref. \cite{Malyuta_2022}.


\subsection{Discretization}

We use a first-order hold interpolation to approximate the continuous problem in Eq.~\ref{eqn:linearized_continuous_form} between successive discrete times in the set $\{t_k\}_{k \in [N_k]}$ that spans the interval $[t_0, t_f]$. The index set $[N_k] \coloneqq \{k \in \mathbb{Z} : 1\leq k \leq N_k\}$, where $N_k$ is the number of discrete time nodes. Later, we describe how these nodes are selected. 

Between the nodes $k$ and $k + 1$, the interpolated continuous time control and the linearized dynamics are
\begin{equation}
    \boldsymbol{u}^j(t) = \lambda_k^{-}(t)\boldsymbol{u}^j_k + \lambda_k^{+}(t)\boldsymbol{u}^j_{k + 1}
\end{equation}
and
\begin{equation}
     \dot{\boldsymbol{x}}^j(t) = A^j(t)\boldsymbol{x}^j(t) + B^j(t)\lambda_k^{-}(t)\boldsymbol{u}^j_k 
                    + B^j(t)\lambda_k^{+}\boldsymbol{u}^j_{k + 1} + \boldsymbol{r}^j(t)\mbox{,}\label{eqn:first_order_hold_continuous_time}
\end{equation}
respectively. The variables $\lambda_k^-(t) = \frac{t_{k + 1} - t}{t_{k + 1} - t_k}$ and $\lambda_k^+(t) = \frac{t  - t_k}{t_{k + 1} - t_k}$. Using this interpolation scheme, the discrete time state transition between node $k$ and $k + 1$ is 
\begin{equation}
    \boldsymbol{x}^j_{k + 1} = A^j_k \boldsymbol{x}^j_k + B_k^{-j} \boldsymbol{u}^j_k + B_k^{+j} \boldsymbol{u}^j_{k + 1} + \boldsymbol{r}^j_k \mbox{,}\\
\end{equation}
where,
\begin{subequations}
    \begin{flalign}
         A^j_k &\coloneqq \Phi^j(t_{k + 1}, t_k)\mbox{,}\\
         B_k^{-j} &\coloneqq A^j_k\int_{t_k}^{t_{k + 1}} \Phi^j(\tau, t_k)^{-1}
         B^j(\tau)\lambda_{k}^-(\tau) d\tau \mbox{,}\\
         B_k^{+j} &\coloneqq A^j_k\int_{t_k}^{t_{k + 1}} \Phi^j(\tau, t_k)^{-1} B^j(\tau) \lambda_{k}^{+}(\tau) d\tau\mbox{,} \\
         \boldsymbol{r}^j_k &\coloneqq A^j_k \int_{t_k}^{t_{k + 1}} \Phi^j(\tau, t_k)^{-1} \boldsymbol{r}^j(\tau) d\tau.
    \end{flalign}
    \label{eqn:discrete_integral_definitions}%
\end{subequations}

A common characteristic of \ac{SCP} algorithms is \textit{artificial infeasibility}. This phenomenon occurs when the linearized path constraints fail to produce a feasible solution, despite the feasibility of the true nonlinear optimal control problem (see Ref. \cite{Malyuta_2022} for further discussion). To address this, \ac{SCvx} introduces a slack variable, termed a \textit{virtual control}, to each discrete-time dynamics constraint. The virtual controls, which we denote as $\{\boldsymbol{\epsilon}^j_k\}_{k = 1}^N$, relax the constraints, ensuring the feasibility of each intermediate convex subproblem. 

Following the previous discussions, we form a local convex optimization subproblem of the form
\begin{subequations}
    \begin{flalign}
        &\hspace{20ex} \min_{\{ \boldsymbol{x}_k^j, \boldsymbol{u}_k^j, \boldsymbol{\epsilon}_k^j \}_{(j, k) \in \mathcal{J}_c \times [N_k]}} L_d \left(\{ \boldsymbol{x}_k^j, \boldsymbol{u}_k^j, \boldsymbol{\epsilon}_k^j \}_{(j, k) \in \mathcal{J}_c \times [N_k]}\right)&\\
        &\hspace{21ex}\mbox{s.t.} \hspace{1ex} \boldsymbol{x}^j_{k + 1} = A^j_k \boldsymbol{x}^j_k + B_k^{-j}\boldsymbol{u}^j_k + B_k^{+j} \boldsymbol{u}^j_{k + 1} + \boldsymbol{r}^j_k + E^j_k \boldsymbol{\epsilon}^j_k\mbox{,}\label{eqn:linearized_discretized_dynamics}&\\
        &\hspace{25ex} \boldsymbol{g}_{\text{ic}}(\boldsymbol{x}^j_0) = \boldsymbol{0}\mbox{,}&\\
        &\hspace{25ex} \boldsymbol{g}_{\text{tc}}(\boldsymbol{x}^j_f) = \boldsymbol{0}\mbox{,}&\\
        &\hspace{25ex} \|\boldsymbol{u}^j_k\|_2 \leq u^j_{max, k}\mbox{,} &\label{eqn:thrust_constraint_in_linear_objective}\\
        &\hspace{25ex} \|\delta \boldsymbol{x}^j_k \|_2 + \| \delta \boldsymbol{u}^j_k \|_2 \leq \eta \hspace{2ex} \text{for } (j, k) \in \mathcal{J}_c \times [N_k]. &
    \end{flalign}
    \label{eqn:general_convex_problem}%
\end{subequations}
The matrix $E^j_k$ is the virtual control gain matrix and defined as 
\begin{equation}
    E^j_k \coloneqq A^j_k \int_{t_k}^{t_{k+1}} \Phi^j(\tau, t_k)^{-1}E(\tau) d\tau \mbox{.}\label{eqn:virtual_control_gain}%
\end{equation}
The matrix $E(\cdot)$ is typically taken to be the identity matrix, and we adopt this in the present study. The function $L_d(\cdot)$ is the linearized discrete-time objective function, further detailed below.


\subsection{Convex Objective Approximation}

We describe the true nonlinear objective and its linearized approximation, both used by the \ac{SCvx} algorithm. Our objective function is a weighted average of the control effort of all active spacecraft and the information gain. Additionally, a penalty term is added to this objective that quantifies the errors incurred by linearization of the dynamics. These linearization errors are expressed through defect vectors  
\begin{equation}
    \boldsymbol{\delta}^j_k = \boldsymbol{x}^j_{k + 1} - \int_{t_k}^{t_{k + 1}} \boldsymbol{f}(\boldsymbol{x}^j(\tau), \boldsymbol{u}^j(\tau), \tau) d \tau\mbox{.}
\end{equation}
To ensure dynamic feasibility, the nonlinear defects must be driven to zero through successive algorithm iterations. The discretized nonlinear objective is then
\begin{flalign}
    J_d\left(\{\boldsymbol{x}_{k}^j, \boldsymbol{u}_{k}^j, \boldsymbol{\delta}_{k}^j \}_{(j, k) \in \mathcal{J}_c\times [N_k]}\right) = -\frac{\alpha_h}{\lambda_h} I(\tilde{\boldsymbol{X}}; \tilde{\boldsymbol{Y}}) + 
    &(1 - \alpha_h)\sum_{k = 1}^{N_k - 1} \left[
    \frac{\Delta t_k}{2} \sum_{j \in \mathcal{J}_c}   \left(||\boldsymbol{u}^j_{k}||_2 + ||\boldsymbol{u}^j_{k + 1}||_2\right) \right] +\label{eqn:scvx_nonlinear_cost}\\ 
    &\gamma \sum_{k = 1}^{N_k - 1}\left[\frac{\Delta t_k}{2} \sum_{j \in \mathcal{J}_c}\left(||\boldsymbol{\delta}^j_k||_1 + ||\boldsymbol{\delta}^j_{k + 1}||_1\right) \right].\nonumber
\end{flalign}
In the above, a trapezoid integration approximation determines the contributions of the total thrust impulse and the defect cost. The time step increment $\Delta t_k = t_{k + 1} - t_k$, and is not necessarily constant. Additionally, note the negative sign distributed to the information gain term, consistent with the problem of minimizing $J_d(\cdot)$ to maximize the information gain. The parameter $\gamma$ is a positive scalar chosen to be sufficiently large to drive the defects to zero. 


Each \ac{SCvx} subproblem generates an approximation of Eq.~\ref{eqn:scvx_nonlinear_cost} that is compatible with a convex program solver. This involves 1) creating a local approximation of the information gain at the iteration's reference decision variables, and 2) substituting the defects with the virtual control variables so that
\begin{flalign}
    L_d\left( \{\boldsymbol{x}_k^j, \boldsymbol{u}_k^j, \boldsymbol{\epsilon}_k^j \}_{(j, k) \in \mathcal{J}_c\times [N_k]} \right) = -\frac{\alpha_h}{\lambda_h} I_{\text{FO}}\left(\tilde{\boldsymbol{X}}; \tilde{\boldsymbol{Y}}|\{\boldsymbol{x}_p^j\}_{j\in \mathcal{J}_c}\right) +
    &(1 - \alpha_h)\sum_{k = 1}^{N_k - 1} \left[
    \frac{\Delta t_k}{2} \sum_{j \in \mathcal{J}_C}   \left(||\boldsymbol{u}^j_{k}||_2 + ||\boldsymbol{u}^j_{k + 1}||_2\right) \right] +\label{eqn:scvx_linearized_cost}\\ 
    &\gamma \sum_{k = 1}^{N - 1}\left[\frac{\Delta t_k}{2} \sum_{j \in \mathcal{J}_c}\left(||\boldsymbol{\epsilon}^j_k||_1 + ||\boldsymbol{\epsilon}^j_{k + 1}||_1\right) \right].\nonumber
\end{flalign}
In the above, $I_{\text{FO}}(\cdot)$ is a first-order Taylor expansion of our mutual information approximation (not to be confused with the Gaussian first-order approximation discussed previously). Because controllable spacecraft are constrained to be passive during an observation arc,  we only require the expansion with respect to the state variables $\{\boldsymbol{x}_p^j \}_{j \in \mathcal{J}_c}$, where, recall, $p$ indexes the time $t_p$ that corresponds to the start of the observation arc. This approximation is given by
\begin{equation}
    I_{\text{FO}}\left(\tilde{\boldsymbol{X}}; \tilde{\boldsymbol{Y}} | \{\boldsymbol{x}_p^j \}_{j \in \mathcal{J}_c}\right) = \bar{I}(\tilde{\boldsymbol{X}}; \tilde{\boldsymbol{Y}}) + 
    \sum_{j \in \mathcal{J}_c} \nabla_{\boldsymbol{x_p^j}} \bar{I}(\tilde{\boldsymbol{X}}; \tilde{\boldsymbol{Y}})
\cdot \left( \boldsymbol{x}_p^j - \bar{\boldsymbol{x}}_p^j \right).\label{eqn:first_order_MI}
\end{equation}
In the above, the mutual information evaluated at the nominal reference states is defined as

\begin{equation}
    \bar{I}(\tilde{\boldsymbol{X}}, \tilde{\boldsymbol{Y}}) = I(\tilde{\boldsymbol{X}}, \tilde{\boldsymbol{Y}})\bigg|_{ \{\bar{\boldsymbol{x}}_p^j \}_{j \in \mathcal{J}_c}},
\end{equation}
along with the first-order derivatives, 
\begin{equation}
    \nabla_{\boldsymbol{x}_p^j} \bar{I}(\tilde{\boldsymbol{X}}; \tilde{\boldsymbol{Y}}) =  \frac{\partial I(\tilde{\boldsymbol{X}}; \tilde{\boldsymbol{Y}})}{\partial \boldsymbol{x}_p^j}\bigg|_{\{\bar{\boldsymbol{x}}_p^j\}_{j \in \mathcal{J}_c}}.
\end{equation}
In practice, we find it feasible to compute the gradients with automatic differentiation or a five-point stencil finite difference. In the case of a Gaussian first-order mutual information approximation, an analytical expression for this derivative exists, which is described in our previous works \cite{Wolf_2023, Wolf_2024b}.

\subsection{Selection of Discrete Time Nodes}


%


%


This work uses a physically-informed method for distributing discrete time nodes, utilizing the generalized Sundman transformation \cite{Szebehely_1969}. Our approach is similar to the method introduced in Ref. \citenum{Leith_2023}. In regions of the observer's trajectory with significant nonlinearity, sparse node placement limits the accuracy of linearization approximations used by \ac{SCvx} to represent the true dynamics. The Sundman transformation provides an efficient solution for step-size regulation, increasing node density in these nonlinear regions. Temporal regularization is achieved through the transformation
\begin{equation}
    dt = r_m^\nu(\tau) d\tau \mbox{,}
\end{equation}
where $r_m(\cdot)$ is the distance of the initial reference guess trajectory from the Moon, and the exponent $\nu$ is a user-defined parameter. When $\nu = 0$ the fictitious time variable $\tau$ and actual time $t$ are equivalent. However, a positive $\nu$ acts to dilate time near perilune -- the portions of the trajectory that experience the greatest nonlinearity. The time-node spacing is computed by numerically integrating the ordinary differential equation,
\begin{equation}
    \frac{d \boldsymbol{z}}{d \tau} = r^\nu_m(\tau) \begin{bmatrix}
        \boldsymbol{f}(\boldsymbol{x}(\tau), \tau)\\
        1
    \end{bmatrix}\mbox{,}%
\end{equation}
where,
\begin{equation}
    \boldsymbol{z}(\tau) = [\boldsymbol{x}^\top(\tau), t]^\top\mbox{.}%
\end{equation}
In practice, a callback is placed in an \ac{ODE} solver to terminate integration when the state variable $t$ reaches a desired time. 

\section{Numerical Results}\label{sec:results}

\subsection{Equations of Motion}

We apply our methods to consider spacecraft operating in the Earth-Moon three-body system. Our approach imposes no restrictions on model fidelity, but for simplicity, we choose to model the dynamical behavior of both the observer and targets using the \ac{CRTBP} \cite{Szebehely_1969}. The \ac{CRTBP} assumes that the two primary bodies, i.e., the Earth and Moon, revolve in perfect circles around a common barycenter. The equations of motion of a third body are modeled in a rotating reference frame where the $x$-axis points from the system barycenter to the Moon, the $z$-axis is oriented along the system's angular momentum vector, and the $y$-axis completes the right-hand. These equations are 
\begin{flalign}
    \ddot{x} &= 2\dot{y} + x - (1 - \mu)\frac{x + \mu}{r_1^3} - \mu \frac{x + \mu - 1}{r_2^3},\nonumber\\
    \ddot{y} &= -2\dot{x} + y - (1 - \mu)\frac{y}{r_1^3} - \mu \frac{y}{r_2^3},\nonumber\\
    \ddot{z} &= -(1 - \mu)\frac{z}{r_1^3} - \mu \frac{z}{r_2^3}.
    \label{eqn:CRTBP_eom}%
\end{flalign}
The quantities $r_1$ and $r_2$ are defined as
\begin{flalign}
    r_1^2 &= (x + \mu)^2 + y^2 + z^2, \nonumber\\
    r_2^2 &= (x + \mu - 1)^2 + y^2 + z^2.
\end{flalign}
\label{eqn:CRTBP_eom}%

The parameter $\mu = \frac{m_2}{m_1 + m_2}$, where $m_1$ is the mass of the Earth, and $m_2$ is the mass of the Moon. As is common practice in the astrodynamics community, we implement the model using normalized units.  The distance unit, $\text{DU}$, is defined by the distance of the Earth to the Moon, and the time unit is $\frac{1}{2\pi} P_{\text{synodic}}$, where $P_{\text{synodic}}$ is the period of the system around the barycenter.

\subsection{Measurement Model}

The present study considers passive optical bearing and bearing rates as the primary measurements for an observer. The azimuth and elevation angles are 
\begin{equation}
    \theta_k = \arctan2(\boldsymbol{e}^y\cdot\boldsymbol{\rho}_k, \boldsymbol{e}^x \cdot \boldsymbol{\rho}_k)
\end{equation}
and,
\begin{equation}
    \phi_k = \sin^{-1}\left( \frac{\boldsymbol{e}^z \cdot \boldsymbol{\rho}_k}{\|\boldsymbol{\rho}_k\|_2} \right),
\end{equation}
respectively. The azimuth and elevation angle rates are 
\begin{equation}
    \dot{\theta}_k = \frac{(\boldsymbol{e}^x \cdot \boldsymbol{\rho}_k)(\boldsymbol{e}^y\cdot\dot{\boldsymbol{\rho}}_k) - (\boldsymbol{e}^y \cdot \boldsymbol{\rho}_k)(\boldsymbol{e}^x \cdot \dot{\boldsymbol{\rho}}_k)}{\boldsymbol{\rho}_k\cdot\boldsymbol{\rho}_k - (\boldsymbol{e}^z\cdot\boldsymbol{\rho}_k)^2}
\end{equation}
and,
\begin{equation}
    \dot{\phi}_k = \frac{\|\boldsymbol{\rho}_k\|_2(\boldsymbol{e}^z \cdot \dot{\boldsymbol{\rho}}_k) - (\boldsymbol{e}^z \cdot \boldsymbol{\rho}_k)(\boldsymbol{\rho}_k\cdot \dot{\boldsymbol{\rho}}_k)/\|\boldsymbol{\rho}_k\|_2}{\|\boldsymbol{\rho}_k\|_2\sqrt{\boldsymbol{\rho}_k\cdot\boldsymbol{\rho}_k - (\boldsymbol{e}^z \cdot\boldsymbol{\rho}_k)^2}}.
\end{equation}
The relative position and velocity vectors in the above expressions are $\boldsymbol{\rho}_k = \boldsymbol{r}_k^{\text{targ}} - \boldsymbol{r}_k^{\text{obs}}$ and $\dot{\boldsymbol{\rho}}_k = \boldsymbol{v}_k^{\text{targ}} - \boldsymbol{v}_k^{\text{obs}}$, respectively. The unit vectors in each direction are defined as $\boldsymbol{e}^x \coloneqq [1, 0, 0]^\top$, $\boldsymbol{e}^y \coloneqq [0, 1, 0]^\top$, and $\boldsymbol{e}^z \coloneqq [0, 0, 1]^\top$.

The measurement vector stacks the angles and angle rates,
\begin{flalign}
    \boldsymbol{y}_k &= \boldsymbol{h}(\boldsymbol{x}^{\text{targ}}_k, \boldsymbol{x}^{\text{obs}}_k) + \boldsymbol{w}_k \\
    &= [\theta_k, \phi_k, \dot{\theta}_k, \dot{\phi}_k]^\top + \boldsymbol{w}_k,
\end{flalign}
where the measurement noise vector is distributed as 
\begin{equation}
    \boldsymbol{w}_k \sim \mathlarger{\mathcal{N}}\left(\boldsymbol{w}_k; \boldsymbol{0}_{4\times 1}, \text{diag}([\sigma_\theta^2, \sigma_{\phi}^2, \sigma_{\dot{\theta}}^2, \sigma_{\dot{\phi}}^2] \right).
\end{equation}

\subsection{Test Case Parameters}

The test case considers inspection of a single target carried out by a single optical observer, both belong to neighboring \acp{DRO}. The orbit states listed are retrieved from the \ac{JPL} periodic orbit database \cite{JPL_SSD_PeriodicOrbits}. As an aside, the methods developed here are broadly formulated to consider multiple targets and multiple observers of various sensing modes. However, in the interest of introducing the cornerstone aspects of our method, we choose to focus this discussion on the present test case and expand on more complicated scenarios in an accompanying letter. With this in mind, the main purpose of the case study is to evaluate 1) the effect of weighing information gain on the characteristics of the observer's optimal trajectory and expected navigation and tracking performance, and 2) how initial observer uncertainty influences the optimal trajectory when weighing information gain. 

%
\begin{table}[h!]
\caption{\label{tab:test_case_orbits} Initial and terminal target states for the observer and target. These parameters correspond to a \ac{DRO} defined in the \ac{CRTBP}.}
\centering
\setlength{\tabcolsep}{2pt}
\begin{tabular}{ccccccccccccccc}
\hline
\hspace{1ex} & $x$ [DU] & $y$ [DU] & $z$ [DU] & $\dot{x}$ [DU/TU] & $\dot{y}$ [DU/TU] & $\dot{z}$ [DU/TU]  \\\hline
$\boldsymbol{x}^{*\text{obs}}_{0}$ & 0.77818583 & 0.0 & 0.0 & 0.0 & 0.5559319 & 0.0 \\
$\boldsymbol{x}^{*\text{targ}}_{0}$ & 0.77800853 & 0.0 & 0.0 & 0.0 & 0.55619061 & 0.0 \\
$\boldsymbol{x}^{*\text{obs}}_{f}$ & 0.77783122 & 0.0 & 0.0 & 0.0 & 0.55644959 & 0.0\\
$\boldsymbol{x}^{*\text{targ}}_{f}$ & 0.77800853 & 0.0 & 0.0 & 0.0 & 0.55619061 & 0.0\\
\hline 
\end{tabular}
\end{table}

The initial and terminal target states for the observer and target are listed in Table \ref{tab:test_case_orbits}. The numerical simulation is conducted over two periods of the observer's reference orbit. The test case parameters are listed in Table \ref{tab:test_case_params}. Parameters relating to the \ac{SCvx} implementation used here are presented in Table \ref{tab:SCvx_params}. 

The software used to carry out the present case study is written in the \textit{Julia} programming language. We use the \ac{DCP} package \texttt{Convex.jl}\cite{convexjl} along with ECOS \cite{Domahidi_2013}, a second-order cone program solver, to solve each convex subproblem. The software can be executed on a personal laptop, and in this study, the results were generated on a Lenovo Thinkpad X1 Carbon running an Intel i7 CPU at 1.80 GHz with 8 GB of RAM. Each subproblem takes approximately eight seconds of solve time, and a solution is generated within 40 iterations for a maximum solve time of about six minutes. 

\begin{table}[h!]
\caption{\label{tab:test_case_params} Scenario parameters for the numerical test case.}
\centering
\begin{tabular}{lcccccc}
\hline
Parameter & Symbol \& Units &Value\\\hline
Initial observer uncertainty per axis (position) & $\sigma_r$ [km] & 10.0 \\
Initial observer uncertainty per axis (velocity) & $\sigma_v$ [m/s] & $0.1$\\
Initial target uncertainty per axis (position) & $\sigma_r$ [km] & 100.0 \\
Initial target uncertainty per axis (velocity) & $\sigma_v$ [km/s] & $1.0$\\
Maximum thrust acceleration & $a_{\text{max}}$ [$\mu$m/$s^2$] & 20.0 \\
Process noise \ac{PSD} per axis & $\sigma_{a}$ [$\mu$m/$s^{3/2}$]  & $1.0$\\
Bearing angle RMS error & $\sigma_{\theta}$, $\sigma_{\phi}$ [arcsec]  & 2.06265\\
Bearing angle rate RMS error  & $\sigma_{\dot{\theta}}$, $\sigma_{\dot{\phi}}$ [arcsec/sec]  & 0.206265\\
Measurement cadence & $f_{\text{meas}}$ [1/day] & 0.3 \\
Reference period & $P$ [TU] & 3.72471679\\
Scenario duration & [TU] & 2 $P$\\
Observation window & $\mathcal{T}_{\text{obs}}$ [TU, TU] & $[P, 3/2P]$\\
\hline
\end{tabular}
\end{table}

\begin{table}[h!]
\caption{\label{tab:SCvx_params} Hyperparameters for \ac{SCvx} used in the present study.}
\centering
\begin{tabular}{lcccccc}
\hline
Parameter & Symbol \& Units & Value\\\hline\
Sundman regularization parameter & $\nu$ & 1.0\\
Objective scale parameter & $\lambda_h$ & 100.0\\
Number of nodes & $N_k$ & 200\\
Virtual control penalty & $\gamma$ & $1\times10^4$\\
Trust region accuracy metrics & $\rho_0$ & 0.0\\
 -- & $\rho_1$ & 0.1\\
 -- & $\rho_2$ & 0.7\\
Trust region growth factor & $\beta_{\text{sh}}$ & 2.0\\
Trust region shrinkage factor & $\beta_{\text{gr}}$ & 2.0\\
Initial trust region & $\eta_{0}$ & 0.1\\
Trust region lower bound & $\eta_{\text{lb}}$ & $1\times10^{-16}$\\
Trust region upper bound & $\eta_{\text{ub}}$ & $100$\\
\hline
\end{tabular}
\end{table}

\subsubsection{Evaluating the Effect of Mutual Information Weight}

\begin{figure}[hbt!]
    \centering
    \includegraphics[width=0.75\textwidth]{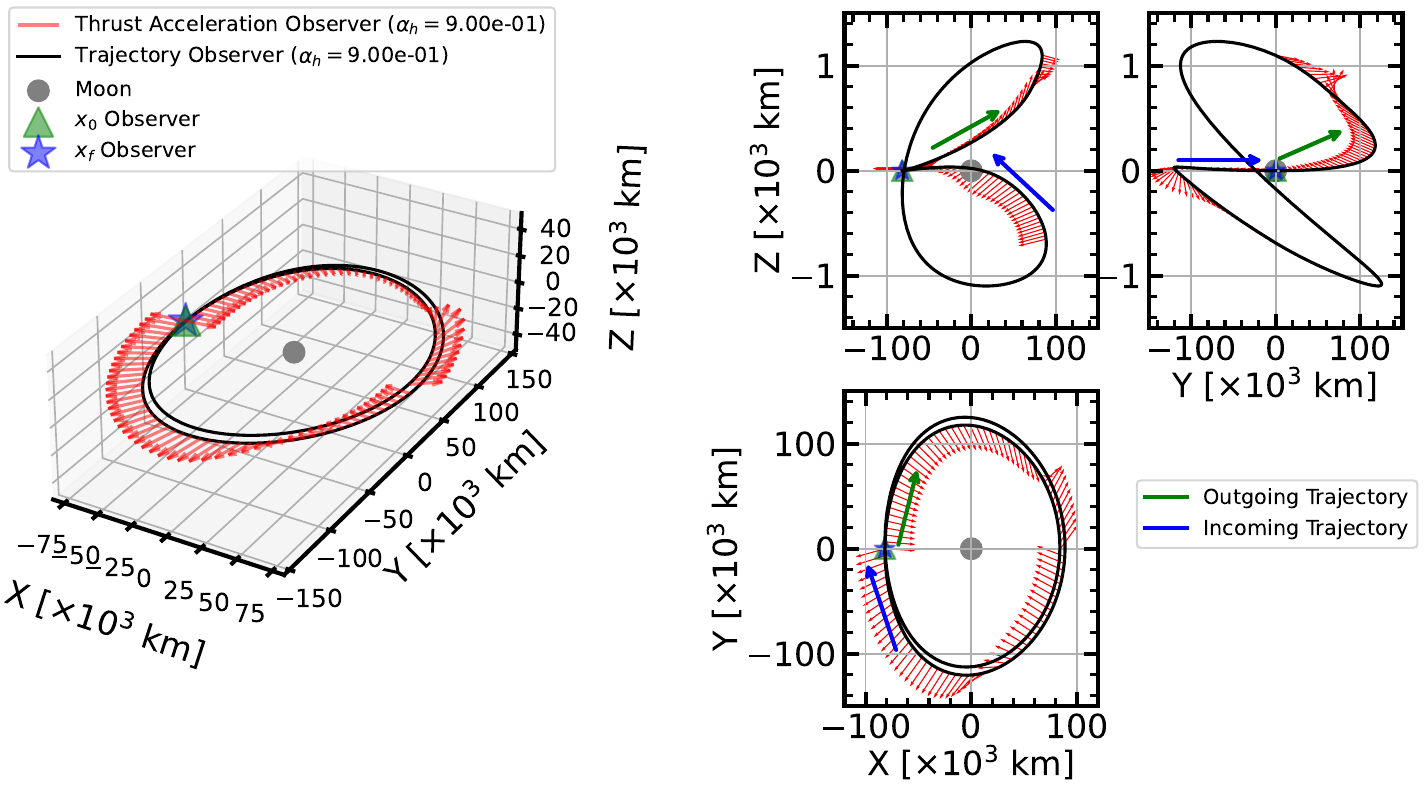}
    \caption{Optimal trajectory of the observer shown in the synodic frame with $\alpha_h = 0.9$. Red quivers indicate thrust magnitude and direction. The green and blue arrows indicate inbound and outbound portions.}\label{fig:absolute_motion}
\end{figure}

We first determine the influence of $\alpha_h$ -- the relative weight of information gain -- on the trajectory characteristics and the expected joint navigation and tracking performance. In the experiment, we increment the scalar weight from $0.0$ to $0.9$. Fully weighing information gain without considering control effort, i.e., $\alpha_h = 1.0$, can sometimes produce chattering behavior in the thrust profile. We posit that it is beneficial to weight control effort in all cases to regularize the problem and produce a useful open-loop control policy. Figure \ref{fig:absolute_motion} plots the absolute motion of the optimal trajectory of the observer for $\alpha_h = 0.9$ in the synodic frame. The left- and right-hand sides show a 3D view, and the trajectory projected onto each coordinate plane, respectively. Red quivers indicate the thrust magnitude and direction of the observer. The green and blue arrows indicate the outbound and inbound portions of the trajectory. In this particular solution example, the observer will thrust during all allowable portions of the scenario, and portions not indicated by the quivers correspond to the observation arc. 

%

%
\begin{figure}[hbt!]
    \centering
    \includegraphics[width=0.75\textwidth]{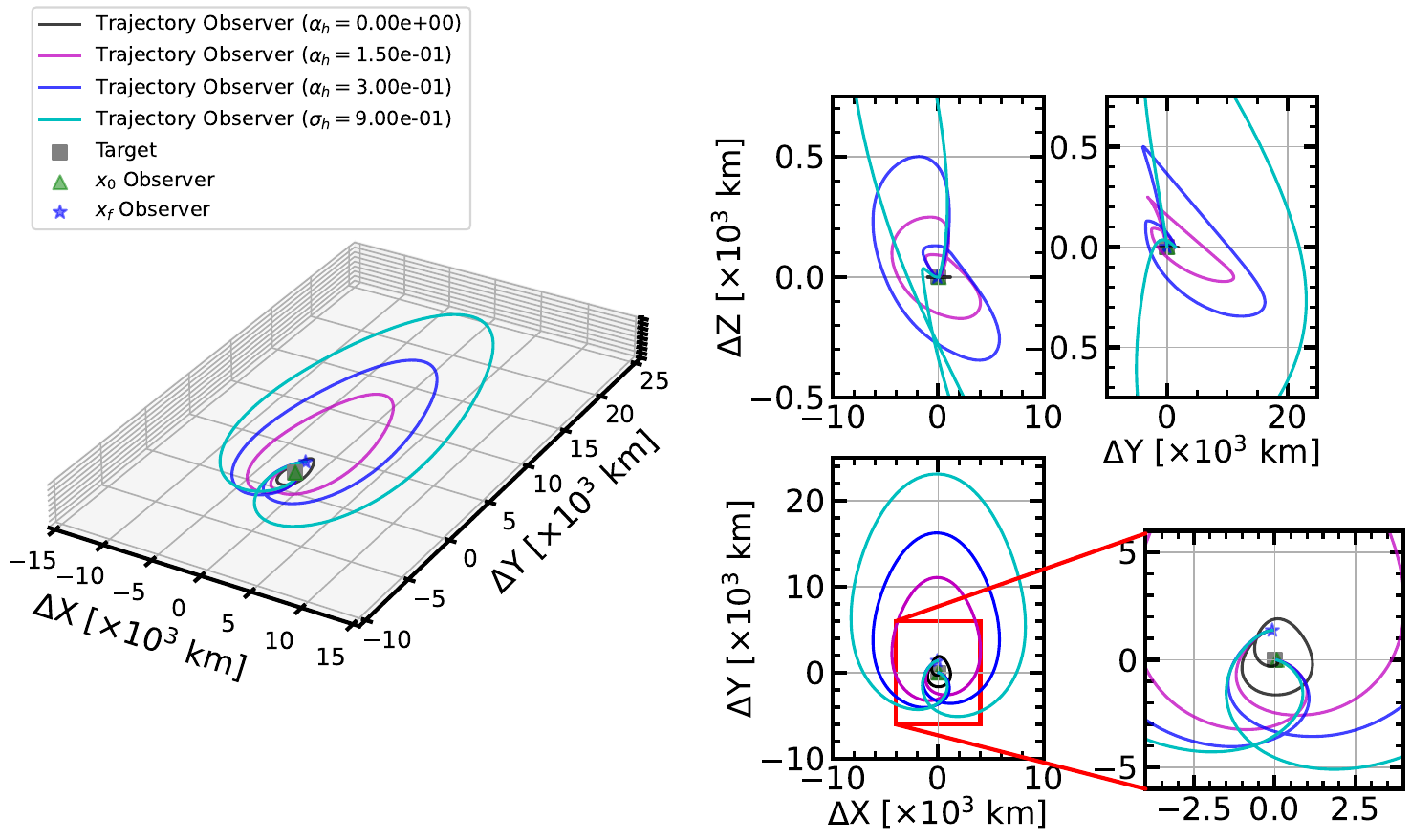}
    \caption{Relative motion of the observer with respect to the target plotted in the synodic frame. Four solutions are shown for different values of $\alpha_h$. In this case, weighing information gain incentivizes the observer to depart from the target's trajectory. }\label{fig:relative_motion}
\end{figure}

The relative motion of the observer around the target is shown in Figure \ref{fig:relative_motion}. Four different optimal solutions are included, corresponding to different relative weighing of information gain. In all cases, the observer starts at and targets the selected states listed in Table \ref{tab:test_case_orbits}. In this example, it is clear that increasing $\alpha_h$ leads the observer to depart further from its target. Most of this motion occurs in the plane, with some out-of-plane motion for higher scalar weighting values. These results suggest that under the conditions listed in Table \ref{tab:test_case_orbits}, improved tracking performance is generated through dynamical differentiation between the target and observer trajectory. As will be shown later, this is not universally true.

%
\begin{figure}[hbt!]
    \centering
    \includegraphics[width=0.75\textwidth]{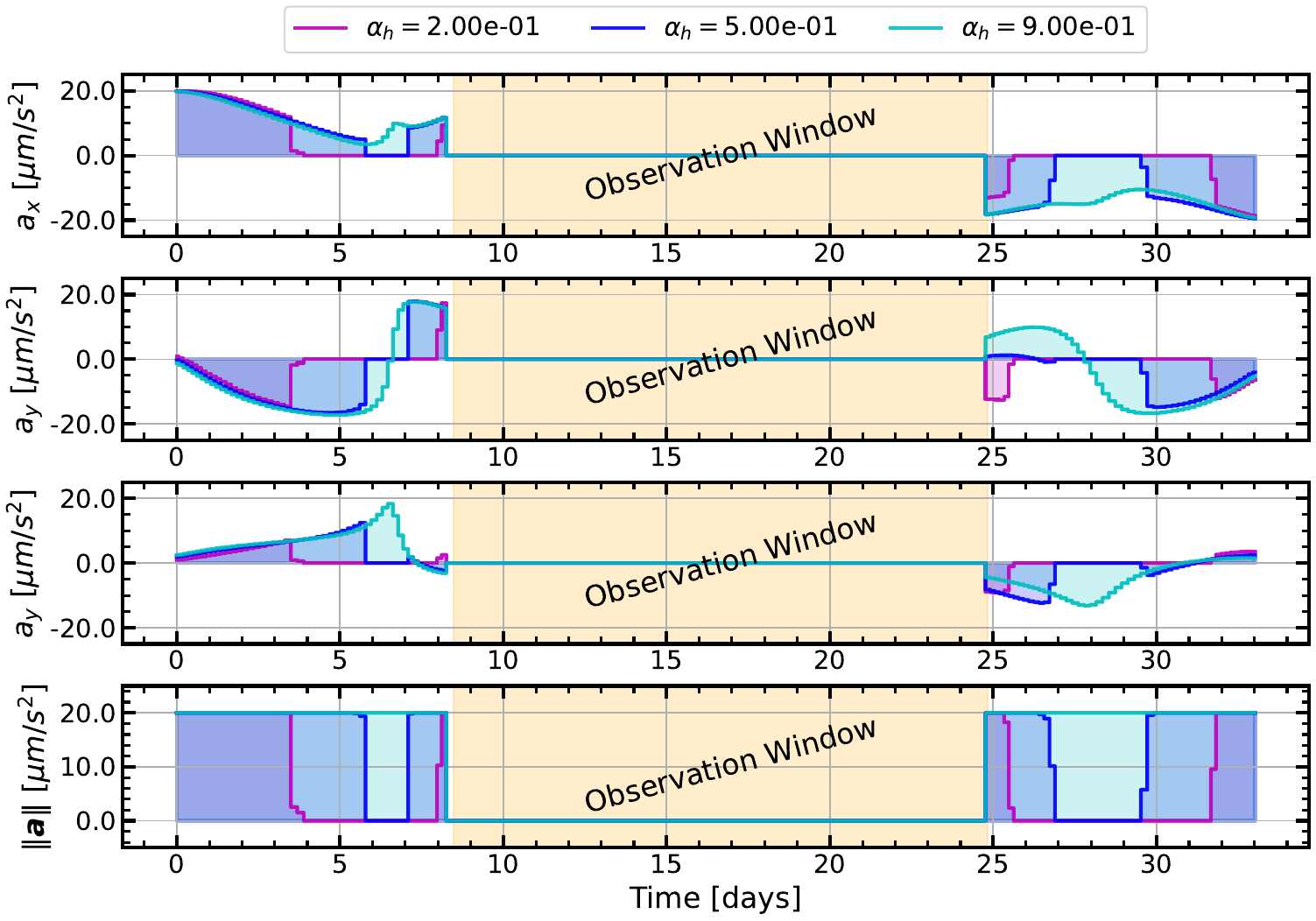}
    \caption{Thrust acceleration profiles produced by solving the optimal control problem with three values of $\alpha_h$.  }\label{fig:thrust_profile}
\end{figure}

%
\begin{figure}[hbt!]
    \centering
    \includegraphics[width=0.75\textwidth]{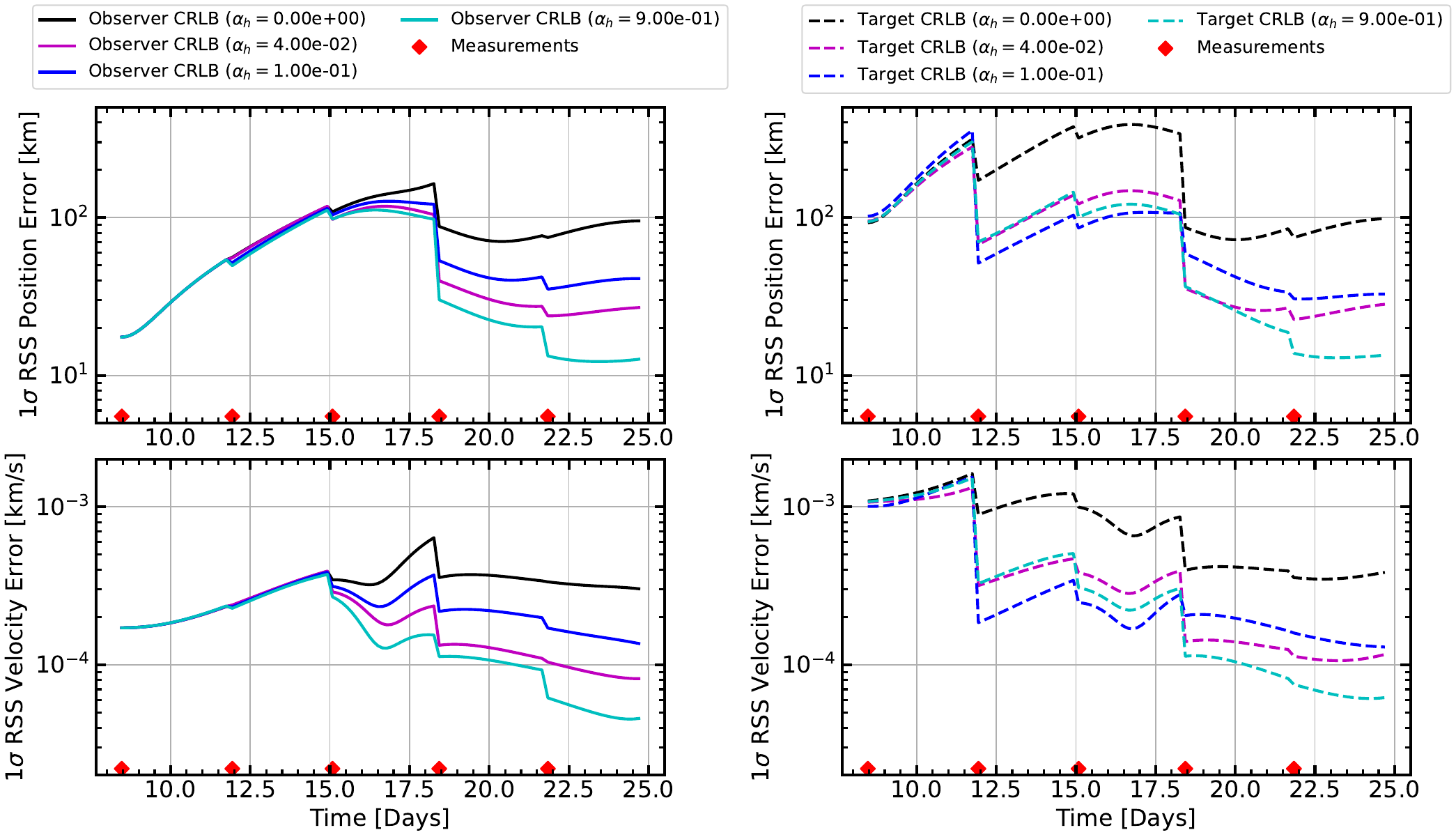}
    \caption{Linear covariance analysis for the observer (left) and target (right). The top and bottom panels plot the expected \ac{RSS} position and velocity errors, respectively. Red diamonds indicated observations.}\label{fig:lin_cov}
\end{figure}

Figure \ref{fig:thrust_profile} plots the thrust acceleration profiles for three information gain weights. The top three show the optimal profiles in each cardinal direction, and the bottom shows the acceleration magnitude. The portion of the scenario between eight and 25 days coincides with the observation interval. During this period, the control is restricted. As expected, increasing $\alpha_h$ uses more propellant. Because control effort is penalized in all cases, it regularizes the problem, and the solutions tend toward bang-on/bang-off profiles typical of minimum-effort optimal control policy. 

We use a linear covariance analysis to assess the impact of information gain on navigation and tracking performance. In our case, this is equivalent to a \ac{CRLB} analysis, which, for nonlinear systems with Gaussian process and measurement noise, is generated through an extended Kalman filter's covariance propagation and update equations, linearized around the true system state \cite{Taylor_1978}. Figure \ref{fig:lin_cov} plots the expected \ac{RSS} position and velocity errors from this analysis for the observer (left), and target (right) over the observation window. The top and bottom panels correspond to position and velocity errors, and we generate these results for four different $\alpha_h$ values. The black curves correspond to the minimum control effort solution, and magenta, blue, and cyan progressively increase  the information gain weight. The red diamonds coincide with optical observations. We find substantial benefit (up to an order of magnitude in position errors) in the expected navigation and tracking performance by including information gain in the trajectory design process. 

\subsubsection{Evaluating the Effect of Observer Uncertainty}

\begin{figure}[hbt!]
    \centering
    \includegraphics[width=0.75\textwidth]{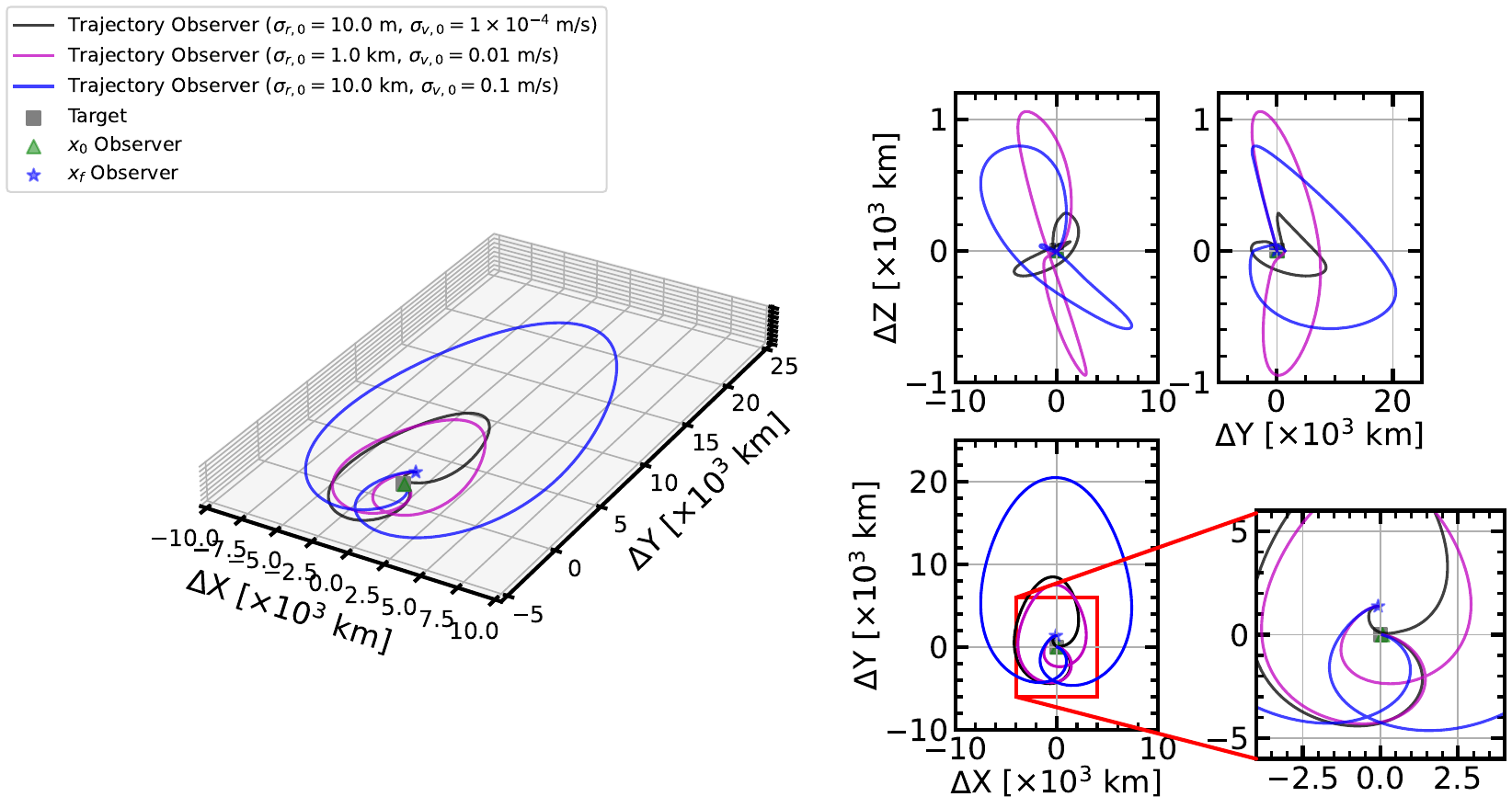}
    \caption{Relative motion of the observer with respect to the target plotted in the synodic frame. Three solutions are shown for different values of $\alpha_h$. In this case, weighing information gain incentivizes the observer to depart from the target's trajectory. }\label{fig:relative_motion_parametric_uncertainty}
\end{figure}

We investigate the influence of the observer's uncertainty on the solution characteristics. In particular, we seek to evaluate the trade-off between dynamical and geometric diversity, and the direct effect of initial uncertainty on these two. To test this, we fix the homotopy at $\alpha_h = 0.5$ -- in between the minimum control and maximum information gain. Three different initial uncertainty levels are considered by scaling the nominal values tabulated in Table \ref{tab:test_case_params} by $10^{-3}$, $0.1$, and $1.0$. In Figure \ref{fig:relative_motion_parametric_uncertainty}, we show the results of this experiment by plotting the relative motion of our observer with respect to the target under these three conditions. There is a clear positive correspondence between the magnitude of the observer's initial uncertainty and the departure from the target space object. Like the previous case, most of the motion is contained in the synodic plane. The results of this test case are consistent with those posited by the experiment described in Section \ref{sec:information_gain}.

%
\section{Conclusions}\label{sec:concluisons}

This study introduces an approach for improving cislunar \ac{SSAT} performance through active maneuvering of agile low-thrust observation platforms. We describe a trajectory optimization framework that jointly considers the observer(s) control expenditure and an information gain functional quantifying the expected navigation and tracking performance. Our framework balances these two competing interests through a simple weighted average -- ideal for a mission designer or automated planner to distill what could otherwise be a complex tradespace to consider. The associated optimal control problem is discretized and solved through \ac{SCvx}. To produce an objective compatible with the proposed scheme, while providing an accurate surrogate of the true information gain, we derive and leverage a Gaussian second-order expansion of the state-augmented time-aggregate  error covariance matrix, which is used to calculate the mutual information. We show that this mutual information approximation behaves as expected in a simple experiment, so that 1) when the observer's state is well constrained, the function leans towards geometric diversity between measurements by operating the observer close to the target, and 2) otherwise favors dynamical diversity by operating a larger baseline from the target to distinguish nonlinearities in the relative motion. Our approach was validated for an \ac{SSAT} scenario with an observer and target belonging to neighboring \acp{DRO}. We demonstrate close to an order of magnitude improvement in the expected navigation and tracking errors, at the expense of control expenditure. 

There are numerous ways to expand the results presented, and here we name a few. As previously mentioned, we intend to elaborate on the versatility of our proposed approach by considering scenarios involving multiple targets, observers, and sensing modes in an accompanying letter. Additionally, we anticipate a fruitful area for future research to be the consideration of joint trajectory planning and sensor scheduling -- that is, managing when and where to maneuver/observe in concert. The methods introduced in the present study could also be leveraged in stochastic planning algorithms to design information-optimal robust spacecraft guidance policies. Finally, practical considerations such as lighting, occultations, and full ephemerides models will be needed to transition the methods produced here into usable algorithms for cislunar missions.  

\section*{Acknowledgments}
The corresponding author thanks Ryan Russell for the useful conversations that improved the quality and presentation of this work. 

\bibliography{references}

\end{document}